\documentclass[10pt,twocolumn,letterpaper]{article}

\usepackage[pagenumbers]{config/style}

\usepackage[pagebackref=true,breaklinks=true,letterpaper=true,colorlinks=true,citecolor=citecolor,urlcolor=magenta,bookmarks=true]{hyperref}

\usepackage{lipsum}

\usepackage[dvipsnames]{xcolor}

\usepackage{color, colortbl}
\usepackage{amsmath}
\usepackage{amsthm}
\usepackage{wrapfig}
\usepackage{array}
\usepackage{newlfont}
\usepackage{textcomp}
\usepackage{multirow}
\usepackage{commath}
\usepackage{hhline}
\usepackage{tabularx}
\usepackage{bigstrut}
\usepackage{afterpage}
\usepackage{algorithmic}
\usepackage{microtype}
\usepackage{mathtools}
\usepackage{booktabs}
\usepackage[ruled]{algorithm2e}
\usepackage{cite}
\usepackage{dingbat}
\usepackage{balance}

\usepackage{times}
\usepackage{epsfig}
\usepackage[export]{adjustbox}
\usepackage{graphicx}
\usepackage{amssymb}
\usepackage{cuted}
\usepackage{capt-of}
\usepackage{float}
\usepackage{enumitem}
\usepackage{comment}
\usepackage{pifont}
\usepackage[toc,page]{appendix}

\usepackage{soul}
\usepackage[usestackEOL]{stackengine}
\usepackage{fancybox}

\newcommand{\ourtitle}{\modelname: \modelnameLong}
\definecolor{citecolor}{HTML}{0071bc}
\definecolor{frontcolor}{HTML}{325ea5}
\definecolor{sidecolor}{HTML}{a58b77}
\definecolor{DeltaColor}{rgb}{0.039,0.73,0.71}
\definecolor{SigmaColor}{rgb}{0.98,0.45,0.0}
\definecolor{AlphaColor}{rgb}{0,0,0.8}
\definecolor{BetaColor}{rgb}{0.8,0,0.8}
\definecolor{GammaColor}{rgb}{0.514,0.34,0.224}
\definecolor{EpsilonColor}{rgb}{0.353,0.725,0.906}
\definecolor{PurpleColor}{HTML}{bca5ea}
\definecolor{OrangeColor}{rgb}{0.914,0.541,0.0.141}
\definecolor{GreenColor}{rgb}{0.137,0.573,0.565}
\definecolor{RedColor}{rgb}{0.949,0.275, 0.224}
\definecolor{LightCyan}{rgb}{0.88,1,1}
\definecolor{Gray}{gray}{0.85}

\newcolumntype{a}{>{\columncolor{Gray}}c}

\newcommand{\highlight}[1]{{\color{black} #1}}
\newcommand{\change}[1]{{\color{black} #1}}

\newcommand{\video}{\textcolor{magenta}{video}\xspace}

\newcommand{\specific}[1]{\xspace{\text{\fontfamily{qcr}\selectfont #1}}\xspace}
\newcommand{\qheading}[1]{\noindent\mbox{\textbf{#1}}}

\newcommand{\colorRef}[1]{\textcolor{black}{#1}}
\usepackage[capitalize]{cleveref}
\crefname{figure}{\colorRef{Fig.}}{\colorRef{Figs.}}
\Crefname{figure}{\colorRef{Figure}}{\colorRef{Figures}}
\crefname{section}{\colorRef{Sec.}}{\colorRef{Secs.}}
\Crefname{section}{\colorRef{Section}}{\colorRef{Sections}}
\Crefname{table}{\colorRef{Table}}{\colorRef{Tables}}
\crefname{table}{\colorRef{Tab.}}{\colorRef{Tabs.}}

\renewcommand{\etc}{\mbox{etc}\xspace}
\renewcommand{\etal}{\mbox{et al.}\xspace}
\renewcommand{\ie}{\mbox{i.e.}\xspace}
\renewcommand{\eg}{\mbox{e.g.}\xspace}

\newcommand{\xmark}{\textcolor{RedColor}{\ding{55}}\xspace}
\newcommand{\cmark}{\textcolor{GreenColor}{\ding{51}}\xspace}
\newcommand{\wcmark}{\textcolor{OrangeColor}{\ding{51}}\xspace}

\newcolumntype{x}[1]{>{\centering\arraybackslash}p{#1pt}}
\newcolumntype{y}[1]{>{\raggedright\arraybackslash}p{#1pt}}
\newcolumntype{z}[1]{>{\raggedleft\arraybackslash}p{#1pt}}
\newlength\savewidth\newcommand\shline{\noalign{\global\savewidth\arrayrulewidth
  \global\arrayrulewidth 1pt}\hline\noalign{\global\arrayrulewidth\savewidth}}

\newcommand{\rgb}{\mbox{RGB}\xspace}
\newcommand{\rgbD}{\mbox{RGB-D}\xspace}

\newcommand{\agora}{\mbox{AGORA}\xspace}
\newcommand{\agoraFIFTY}{\mbox{``AGORA-50''}\xspace}

\newcommand{\threeppl}{\mbox{3DPeople}\xspace}
\newcommand{\humanalloy}{\mbox{Humanalloy}\xspace}

\newcommand{\pymaf}{\mbox{PyMAF}\xspace}

\newcommand{\renderppl}{\mbox{Renderpeople}\xspace}
\newcommand{\thuman}{\mbox{THuman}\xspace}
\newcommand{\axyz}{\mbox{AXYZ}\xspace}
\newcommand{\buff}{\mbox{BUFF}\xspace}
\newcommand{\cape}{\mbox{CAPE}\xspace}
\newcommand{\capeFP}{\mbox{``CAPE-FP''}\xspace}
\newcommand{\capeNFP}{\mbox{``CAPE-NFP''}\xspace}

\newcommand{\twindom}{\mbox{Twindom}\xspace}

\newcommand{\clothplus}{\mbox{CLOTH3D++}\xspace}

\newcommand{\smpl}{\mbox{SMPL}\xspace}
\newcommand{\smplx}{\mbox{SMPL-X}\xspace}
\newcommand{\smplANDx}{\mbox{SMPL(-X)}\xspace}

\newcommand{\smplify}{\mbox{SMPLify}\xspace}

\newcommand{\pifuSIM}{\mbox{$\text{PIFu}^*$}\xspace}
\newcommand{\pamirSIM}{\mbox{$\text{PaMIR}^*$}\xspace}
\newcommand{\pifu}{\mbox{PIFu}\xspace}
\newcommand{\monoport}{\mbox{MonoPort}\xspace}
\newcommand{\pifuhd}{\mbox{PIFuHD}\xspace}
\newcommand{\pamir}{\mbox{PaMIR}\xspace}
\newcommand{\arch}{\mbox{ARCH}\xspace}
\newcommand{\archplus}{\mbox{ARCH++}\xspace}

\newcommand{\scanimate}{\mbox{SCANimate}\xspace}

\newcommand{\modelname}{ICON\xspace}

\newcommand{\modelnameLong}{Implicit Clothed humans Obtained from Normals\xspace}
\newcommand{\projectURL}{\url{https://icon.is.tue.mpg.de}}

\newcommand{\groundtruth}{{ground-truth}\xspace}

\newcommand{\stateoftheart}{\mbox{state-of-the-art}\xspace}
\newcommand{\sota}{\mbox{SOTA}\xspace}
\newcommand{\offtheshell}{\mbox{off-the-shell}\xspace}

\newcommand{\inthewild}{\mbox{in-the-wild}\xspace}
\newcommand{\itw}{\inthewild}
\newcommand{\ood}{\mbox{out-of-distribution}\xspace}
\newcommand{\oof}{\mbox{out-of-frame}\xspace}
\newcommand{\twoD}{2D\xspace}
\newcommand{\threeD}{3D\xspace}
\newcommand{\suppl}{\textcolor{magenta}{\textbf{Appx}}\xspace}

\newcommand{\bodyMesh}{\mathcal{M}}
\newcommand{\diffrender}{\mathcal{DR}}
\newcommand{\bodyNormImgFB}{\mathcal{N}^{\text{b}}}
\newcommand{\bodyNormImg}{\mathcal{N}^{\text{b}}}
\newcommand{\normNet}{\mathcal{G}^\text{N}}
\newcommand{\predCloNormImgFB}{\widehat{\mathcal{N}}^{\text{c}}}
\newcommand{\predCloNormImg}{\widehat{\mathcal{N}}^{\text{c}}}
\newcommand{\gtCloNormImgFB}{\mathcal{N}^{\text{c}}}
\newcommand{\pointFeat}{\mathcal{F}_\text{P}}
\newcommand{\cloNormFeat}{\mathcal{F}_\text{n}^\text{c}}
\newcommand{\bodyNormFeat}{\mathcal{F}_\text{n}^\text{b}}
\newcommand{\bodySDF}{\mathcal{F}_\text{s}}

\newcommand{\imFunc}{\mathcal{IF}}
\newcommand{\body}{\text{b}}
\newcommand{\cloth}{\text{c}}
\newcommand{\normal}{\text{N}}

\usepackage{acronym}
\acrodef{amt}[AMT]{Amazon Mechanical Turk}

\begin{document}

\title{\ourtitle}

\author{
Yuliang Xiu \quad Jinlong Yang \quad Dimitrios Tzionas \quad Michael J. Black\\
Max Planck Institute for Intelligent Systems, T{\"u}bingen, Germany\\
{\tt\small \{yuliang.xiu, jinlong.yang, dtzionas, black\}@tuebingen.mpg.de}\\
}

\newcommand{\teaserCaption}{
{\bf Images to avatars.}
\modelname robustly reconstructs 3D clothed humans in unconstrained poses from individual video frames (Left). 
These are used to learn a fully textured and animatable clothed avatar with realistic clothing deformations (Right).
}

\twocolumn[{
    \renewcommand\twocolumn[1][]{#1}
    \maketitle
    \centering
    \vspace{-0.5em}
    \begin{minipage}{1.00\textwidth}
        \centering
        \includegraphics[trim=000mm 000mm 000mm 000mm, clip=False, width=\linewidth]{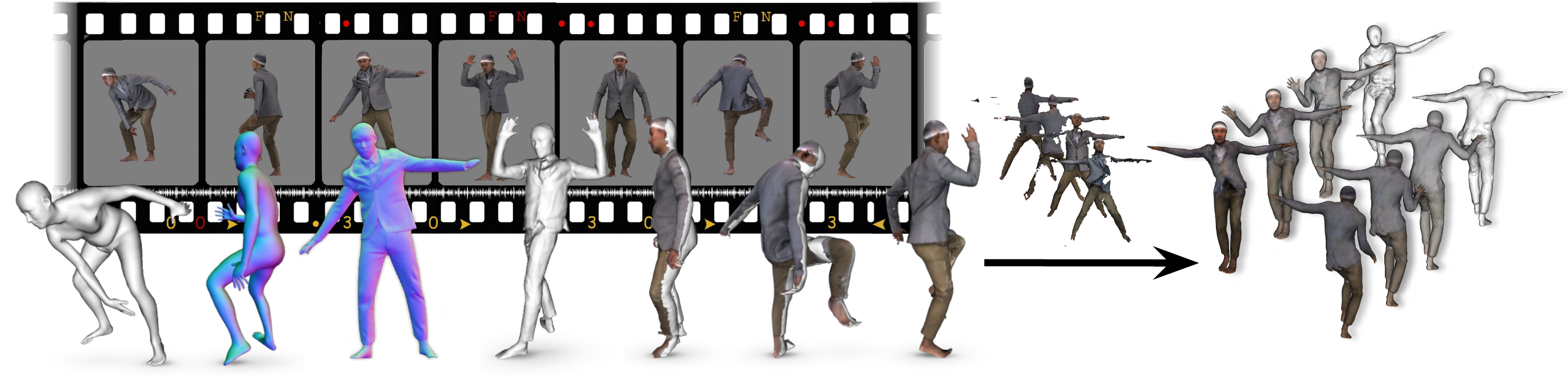}
    \end{minipage}
    \vspace{-0.5 em}
    \captionof{figure}{\teaserCaption}
    \label{fig:teaser}
    \vspace{2.2em}
}]

\begin{abstract}
\vspace{-1.0em}
Current methods for learning realistic and animatable \threeD clothed avatars need either posed \threeD scans or \twoD images with carefully controlled user poses. 
In contrast, our goal is to learn an avatar from only \twoD images of people in \textsl{unconstrained}  poses. 
Given a set of images, our method estimates a detailed \threeD surface from each image and then combines these into an animatable avatar. 
Implicit functions are well suited to the first task, as they can capture details like hair and clothes. 
Current methods, however, are not robust to varied human poses and often produce \threeD surfaces with broken or disembodied limbs, missing details, or non-human shapes. 
The problem is that these methods use global feature encoders that are sensitive to global pose. 
To address this, we propose \modelname (``\modelnameLong''), which, instead, uses local features. 
\modelname has two main modules, both of which exploit the \smplANDx body model. 
First, \modelname infers detailed clothed-human normals (front/back) conditioned on the \smplANDx normals. 
Second, a visibility-aware implicit surface regressor produces an iso-surface of a human occupancy field. 
Importantly, at inference time, a feedback loop alternates between refining the \smplANDx mesh using the inferred clothed normals and then refining the normals. 
Given multiple reconstructed frames of a subject in varied poses, we use a modified version of \scanimate to produce an animatable avatar from them. 
Evaluation on the \agora and \cape datasets shows that \modelname outperforms the state of the art in reconstruction, even with heavily limited training data. 
Additionally, it is much more robust to \ood samples, \eg, \inthewild poses/images and out-of-frame cropping. 
\modelname takes a step towards robust \threeD clothed human reconstruction from \inthewild images. 
This enables avatar creation directly from video with personalized pose-dependent cloth deformation. 
Models and code are available for research at \projectURL.
\end{abstract}

\vspace{-1.0em}
\section{Introduction}

Realistic virtual humans will play a central role in mixed and augmented reality, forming a key foundation for the ``metaverse'' and supporting remote presence, collaboration, education, and entertainment.
To enable this, new tools are needed to easily create \threeD virtual humans that can be readily animated.
Traditionally, this requires significant artist effort and expensive scanning equipment.
Therefore, such approaches do not scale easily.
A more practical approach would enable individuals to create an avatar from one or more images.
There are now several methods that take a single image and regress a minimally clothed \threeD human model \cite{yao2021pixie,pavlakos2019expressive,kolotouros2019spin,alldieck2018videoavatar,alldieck2019tex2shape, SHAPY_2022}.
Existing parametric body models, however, lack important details like clothing and hair  \cite{Joo2018_adam,SMPL:2015,pavlakos2019expressive,romero2017embodied,xu2020ghum}.
In contrast, we present a method that robustly extracts \threeD scan-like data from images of people in arbitrary poses and uses this to construct an animatable avatar.

\begin{figure}
	\centering
    \vspace{-0.5 em}
    \includegraphics[trim=000mm 000mm 000mm 000mm, clip=true, width=1.00 \linewidth]{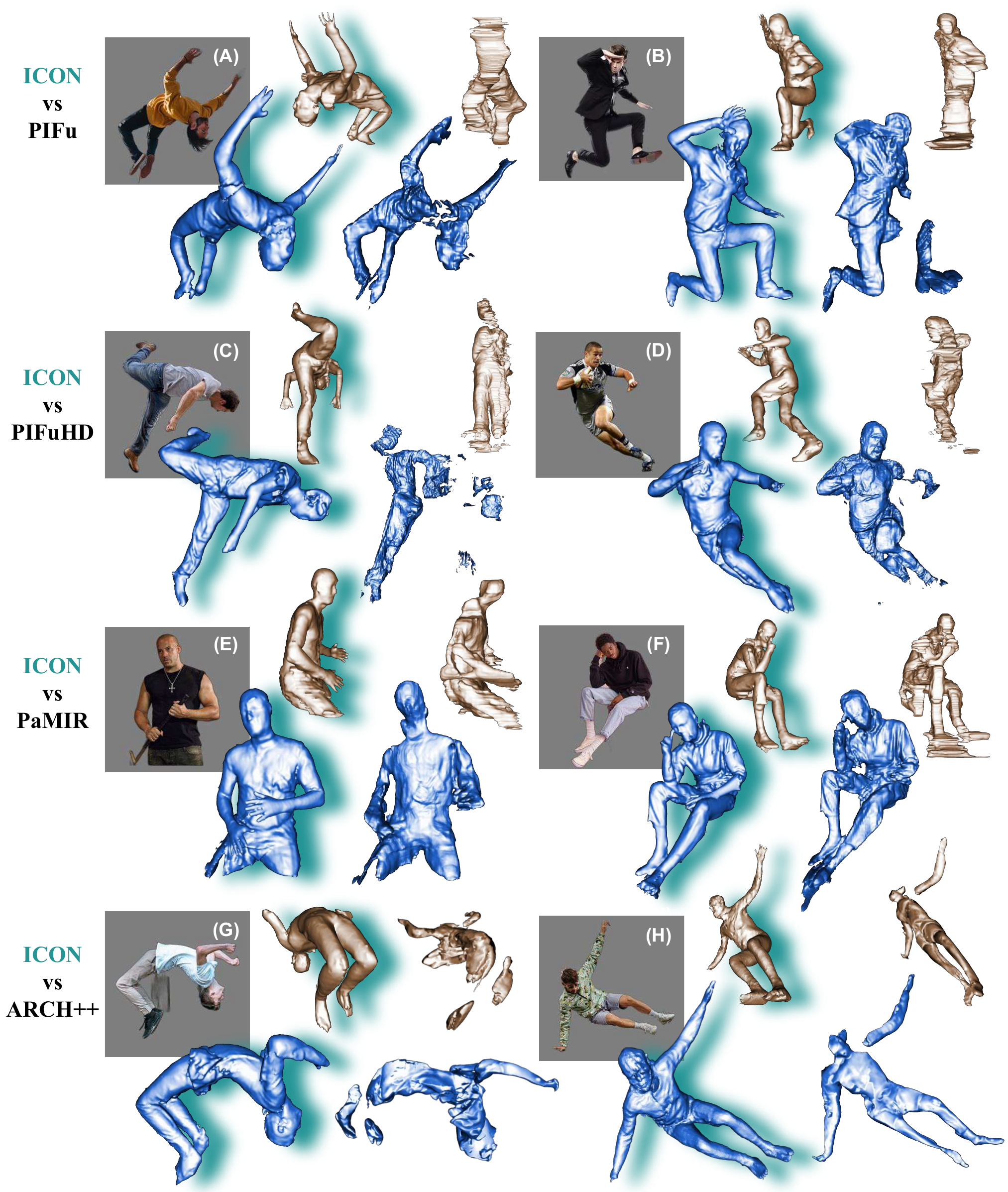}
    \vspace{-2.0 em}
	\caption{
	           \highlight{SOTA methods for inferring \threeD humans} from \inthewild images, \eg, \pifu, \pifuhd, \pamir, and \archplus, \highlight{struggle with}
	           challenging poses and out-of-frame cropping (E), resulting in various artifacts including non-human shapes (A,G), disembodied parts (B,H), missing body parts (C,D), missing details (E), and high-frequency noise (F).
	           %
	           %
	           \modelname deals with these challenges
	           and produces high-quality results, 
	           highlighted with a \textcolor{GreenColor}{green shadow}
	           \highlight{Front view (\textcolor{frontcolor}{blue}) and rotated view (\textcolor{sidecolor}{bronze}).}
	}
	\label{fig:motivation}
\end{figure}

We base our approach on implicit functions \highlight{(IFs)}, which go beyond parametric body models to represent fine shape details and varied topology.
%
%
\highlight{IFs allow recent methods} 
to infer detailed shape from an image \cite{saito2019pifu,saito2020pifuhd,zheng2020pamir,huang2020arch,he2020geoPifu,yang2021s3}.
Despite promising results, \stateoftheart~\highlight{(\sota)} methods struggle with \inthewild data and often produce 
humans with broken or disembodied 
\highlight{limbs}, missing details, high-frequency noise, or non-human shape; see \cref{fig:motivation} for examples.

The issues with previous methods are twofold:
\mbox{(1)   Such methods are} typically trained on small, hand-curated, \threeD human datasets (\eg~\renderppl~\cite{renderpeople}) with very limited pose, shape and clothing variation. 
\mbox{(2)   They typically feed} their implicit-function module with features of a global \twoD image 
            or \threeD voxel encoder, but these are sensitive to global pose. 
While more, and more varied, \threeD training data would help, such data remains limited.
Hence, we take a different approach and improve the model.

Specifically, our goal is to reconstruct a detailed clothed \threeD human from a single \rgb image with a method that is training-data efficient and robust to \inthewild images and \ood~poses.
Our method, 
\highlight{called} \emph{\modelname}, stands for \emph{\modelnameLong}. 
\modelname replaces the global encoder of existing methods with a more data-efficient local scheme; \cref{fig:architecture} shows a model overview. 
\modelname takes as input an \rgb image of a segmented clothed human and a \smpl body estimated from the image~\cite{Kocabas2021pare}.
The \smpl body is used to guide two of \modelname's modules: 
one infers detailed clothed-human surface normals (front and back views), and 
the other infers a visibility-aware implicit surface (iso-surface of an occupancy field).
Errors in the initial \smpl estimate, however, might misguide inference. 
Thus, at inference time, an iterative feedback loop refines \smpl (\ie, its \threeD shape, pose, and translation) using the inferred detailed normals, and vice versa, leading to a refined implicit shape with better \threeD details.

We evaluate \modelname quantitatively and qualitatively on challenging datasets, namely \agora~\cite{patel2021agora} and \cape~\cite{ma2020cape}, as well as on \inthewild images. 
Results show that \modelname has two advantages \wrt the state of the art: 
\qheading{(1) Generalization}. 
\modelname's locality helps it generalize to \inthewild images and \ood poses and clothes better than previous methods.
Representative cases are shown in \cref{fig:motivation}; notice that, although \modelname is trained on full-body images only, it can handle 
images with out-of-frame cropping, with no fine tuning or post processing.
\qheading{(2) Data efficacy}. 
\modelname's locality 
\highlight{helps it avoid} 
spurious correlations between pose and surface shape.
Thus, it \highlight{needs} 
less data for training.
\modelname significantly outperforms baselines in low-data regimes, as it reaches 
\highlight{\sota}
performance when trained 
with as little as $12\%$ of the data.

We provide an example application of \modelname for creating an animatable avatar; see \cref{fig:teaser} for an overview. 
We first apply \modelname on the individual frames of a video sequence, to obtain \threeD meshes of a clothed person in various poses. 
We then use these to train a poseable avatar using a modified version of \scanimate~\cite{saito2021scanimate}.
Unlike \threeD scans, which \scanimate takes as input, our estimated shapes are not equally detailed and reliable from all views.
Consequently, we modify \scanimate to exploit visibility information in learning the avatar.
The output is a \threeD clothed avatar that moves and deforms naturally; see \cref{fig:teaser}-right and \cref{fig:scanimate}.

\modelname takes a step towards robust reconstruction of \threeD clothed humans from \inthewild photos. 
Based on this, fully textured and animatable avatars with personalized pose-aware clothing deformation can be created directly from video frames.
Models and code are available at \projectURL.

\section{Related work}

\textbf{Mesh-based statistical models.} 
Mesh-based statistical body models~\cite{Joo2018_adam,SMPL:2015,pavlakos2019expressive,romero2017embodied,xu2020ghum} are a popular explicit representation for \threeD human reconstruction. 
This is not only because such models capture the statistics across a human population, but also because meshes are compatible with standard graphics pipelines.
A lot of work~\cite{kanazawa2018hmr,kolotouros2019spin,kocabas2020vibe,Yi_MOVER_2022,sun2021BEV,smith2019facsimile,SHAPY_2022} estimates \threeD body meshes from an \rgb image, but these have no clothing. 
Other work estimates clothed humans, instead, by modeling clothing geometry as \threeD offsets on top of body geometry \cite{alldieck2018videoavatar,alldieck2018dv,alldieck2019peopleInClothing,zhu2019hierarchMeshDeform,lazova2019textures360,ponsMoll2017clothCap,alldieck2019tex2shape,xiang2020monoClothCap}. 
The resulting clothed \threeD humans can be easily animated, as they naturally inherit the skeleton and surface skinning weights from the underlying body model. 
An important limitation, though, is modeling clothing such as skirts and dresses; 
since these differ a lot from the body surface, simple body-to-cloth offsets are insufficient. 
To address this, some methods~\cite{bhatnagar2019multiGarmentNet,jiang2020bcnet} use a classifier to identify cloth types in the input image, and then perform cloth-aware inference for \threeD reconstruction. 
However, such a remedy does not scale up to a large variety of clothing types. 
Another advantage of mesh-based statistical models, is that texture information can be easily accumulated through multi-view images or image sequences~\cite{alldieck2018videoavatar,bhatnagar2019multiGarmentNet}, due to their consistent mesh topology. 
The biggest limitation, though, is that the state of the art does not generalize well \wrt clothing-type variation, and it estimates meshes that do not align well to input-image pixels.

\textbf{Deep implicit functions.} 
Unlike meshes, deep implicit functions \cite{park2019deepSDF,mescheder2019occNet,chen2018implicit_decoder} can represent detailed \threeD shapes with arbitrary topology, and have no resolution limitations. 
Saito \etal~\cite{saito2019pifu} introduce deep implicit functions for clothed \threeD human reconstruction from \rgb images and, later \cite{saito2020pifuhd}, they significantly improve \threeD geometric details. 
The estimated shapes align well to image pixels. 
However, their shape reconstruction lacks regularization, and often produces artifacts like broken or disembodied limbs, missing details, or geometric noise. 
He \etal~\cite{he2020geoPifu} add a coarse-occupancy prediction branch, and Li \etal~\cite{Li2020portrait} and Dong \etal~\cite{dong2022pina} use depth information captured by an \rgbD camera to further regularize shape estimation and provide robustness to pose variation. 
Li \etal~\cite{li2020realtimeVolMoCap,li2020monoportRTL} speed up inference through an efficient volumetric sampling scheme. 
A limitation of all above methods is that the estimated \threeD humans cannot be reposed,
because implicit shapes (unlike statistical models) lack a consistent mesh topology, a skeleton, and skinning weights. 
To address this, Bozic \etal~\cite{bozic2020neural} infer an embedded deformation graph to manipulate implicit functions, while Yang \etal~\cite{yang2021s3} also infer a skeleton and skinning fields.

\textbf{Statistical models \& implicit functions.}
Mesh-based statistical models are well regularized, while deep implicit functions are much more expressive.
To get the best of both worlds, recent methods \cite{bhatnagar2020implicitParModel,huang2020arch,bhatnagar2020loopReg,zheng2020pamir} combine the two representations. 
Given a sparse point cloud of a clothed person, IPNet~\cite{bhatnagar2020implicitParModel} infers an occupancy field with body/clothing layers, registers \smpl to the body layer with inferred body-part segmentation, and captures clothing as offsets from \smpl to the point cloud.
Given an \rgb image of a clothed person, ARCH~\cite{huang2020arch} and ARCH++~\cite{he2021ICCVarchplus} 
reconstruct \threeD human shape in a canonical space by warping query points from the canonical to the posed space, and projecting them onto the \twoD image space. 
However, to train these models, one needs to unpose scans into the canonical pose with an accurately fitted body model; inaccurate poses cause artifacts.
Moreover, unposing clothed scans using the ``undressed'' model's skinning weights alters shape details. 
For the same \rgb input, Zheng \etal~\cite{zheng2020pamir,zheng2021deepmulticap} condition the implicit function on a posed and voxelized \smpl mesh for robustness to pose variation and reconstruct local details from the image pixels, similar to \pifu~\cite{saito2019pifu}. 
However, these methods are sensitive to global pose, due to their \threeD convolutional encoder. 
Thus, for training data with limited pose variation, they struggle with \ood poses and \itw images.

\textbf{Positioning \modelname~\wrt related work.}
\modelname combines the statistical body model \smpl with an implicit function, to reconstruct clothed \threeD human shape from a single \rgb image. 
\smpl not only guides \modelname's estimation, but is also optimized ``in the loop" during inference to enhance its pose accuracy. 
Instead of relying on the global body features, \modelname exploits local body features that are agnostic to global pose variations. 
As a result, even when trained on heavily limited data,  \modelname achieves \stateoftheart performance and is robust to \ood poses. 
This work links monocular \threeD clothed human reconstruction to scan/depth based avatar modeling algorithms \cite{saito2021scanimate,MetaAvatar2021Wang,chen2021snarf,tiwari21neuralgif,deng2019nasa,Ma:CVPR:2021,POP:ICCV:2021}.

\section{Method}
\label{sec: method}

\begin{figure*}
    \centerline{
    \includegraphics[trim=000mm 020mm 000mm 0011mm, clip=True, width=1.0\textwidth]{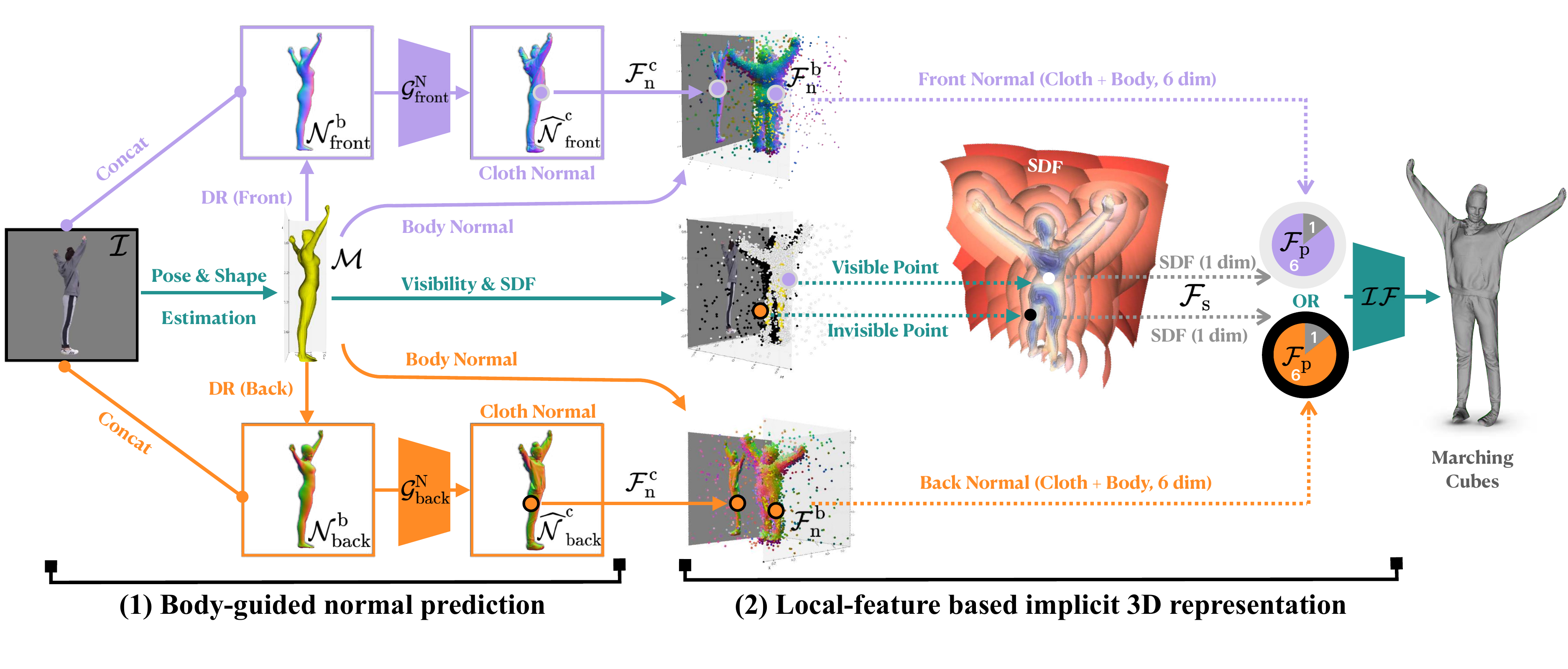}}
    \vspace{-0.5 em}
    \caption{
                \modelname's architecture contains two main modules for: 
                (1) body-guided normal prediction, and 
                (2) local-feature based implicit \threeD reconstruction.
                The dotted line with an arrow is a \twoD or \threeD query function.
                The two $\normNet$ networks (\textcolor{PurpleColor}{purple}/\textcolor{OrangeColor}{orange}) 
                have different parameters.
    }
    \label{fig:architecture}
    \vspace{-1.0em}
\end{figure*}

\modelname is a deep-learning model that infers a \threeD clothed human from a color image. 
Specifically, \modelname takes as input an \rgb image with a segmented clothed human 
\highlight{(following the suggestion of
\href{https://github.com/facebookresearch/pifuhd\#testing}{\pifuhd}'s repository \cite{PIFuHD_code})}, along with an estimated human body shape ``under clothing'' (\smpl), and outputs a pixel-aligned \threeD shape reconstruction of the clothed human. 
\modelname has two main modules (see \cref{fig:architecture}) for: (1) \smpl-guided clothed-body normal prediction and (2) local-feature based implicit surface reconstruction.

\subsection{Body-guided normal prediction}
\label{sec: smpl-conditioned normal predict}

Inferring full-$360^{\circ}$ \threeD normals from a single \rgb image of a clothed person is challenging; normals for the occluded parts need to be hallucinated based on the observed parts. 
This is an ill-posed task and is challenging for deep networks. 
\highlight{Unlike model-free \highlight{methods}~\cite{Jafarian_2021_CVPR_TikTok, saito2020pifuhd, tang2019aneural}, }
\modelname takes into account a \smpl~\cite{SMPL:2015} ``body-under-clothing'' mesh to reduce ambiguities and guide front and (especially) back clothed\highlight{-body} normal prediction. 
To estimate the \smpl mesh  $\bodyMesh(\beta, \theta) \in \mathbb{R}^{N \times 3}$ from image $\mathcal{I}$, we use \highlight{\pymaf~\cite{zhang2021pymaf} due to its 
better mesh-to-image alignment compared to other methods}. 
\smpl is parameterized by shape, $\beta \in \mathbb{R}^{10}$, and pose, $\theta \in \mathbb{R}^{3 \times K}$, where $N = 6,890$ vertices and $K = 24$ joints. 
\modelname is also compatible with \smplx~\cite{pavlakos2019expressive}.

Under a weak-perspective camera model, with scale $s \in \mathbb{R}$ and translation $t \in \mathbb{R}^3$, we use the PyTorch3D~\cite{ravi2020pytorch3d} differentiable renderer, denoted as $\diffrender$, to render $\bodyMesh$ from two opposite views, obtaining ``front'' (\ie, observable side) and ``back'' (\ie, occluded side) \smpl-body normal maps $\bodyNormImgFB = \{\bodyNormImg_\text{front}, \bodyNormImg_\text{back}\}$. 
Given $\bodyNormImgFB$ and the original color image $\mathcal{I}$, our normal networks $\normNet=\{\normNet_\text{front}, \normNet_\text{back}\}$  predict clothed-body normal maps, denoted as $\predCloNormImgFB = \{\predCloNormImgFB_\text{front}, \predCloNormImgFB_\text{back}\}$:
\begin{align}
         \diffrender(\bodyMesh) &\rightarrow \bodyNormImgFB \text{,}\\
         \normNet(\bodyNormImgFB, \mathcal{I}) &\rightarrow{\predCloNormImgFB} \text{.}
         \label{eq:normal-infer}
\end{align}
We train the normal networks, $\normNet$, with the following loss:
\begin{align}
 \label{eq:normal-loss}
        \mathcal{L}_\normal      &= \mathcal{L}_\text{pixel} + \lambda_\text{VGG} \mathcal{L}_\text{VGG} \text{,}
\end{align}
where 
$\mathcal{L}_\text{pixel} = | \gtCloNormImgFB_\text{v} - \predCloNormImgFB_\text{v} |$, $\text{v} = \{\text{front}, \text{back}\}$, is a loss (L1) between 
\groundtruth and predicted normals (the two $\normNet$ in \cref{fig:architecture} have different parameters), 
and $\mathcal{L}_\text{VGG}$ is a perceptual loss \cite{justin2016perceploss} weighted by $\lambda_\text{VGG}$.  
With only $\mathcal{L}_\text{pixel}$, the inferred normals are blurry, but adding $\mathcal{L}_\text{VGG}$ helps recover details.

\begin{figure}
    \centering
    \includegraphics[trim=000mm 000mm 000mm 000mm, clip=true, width=1.00\linewidth]{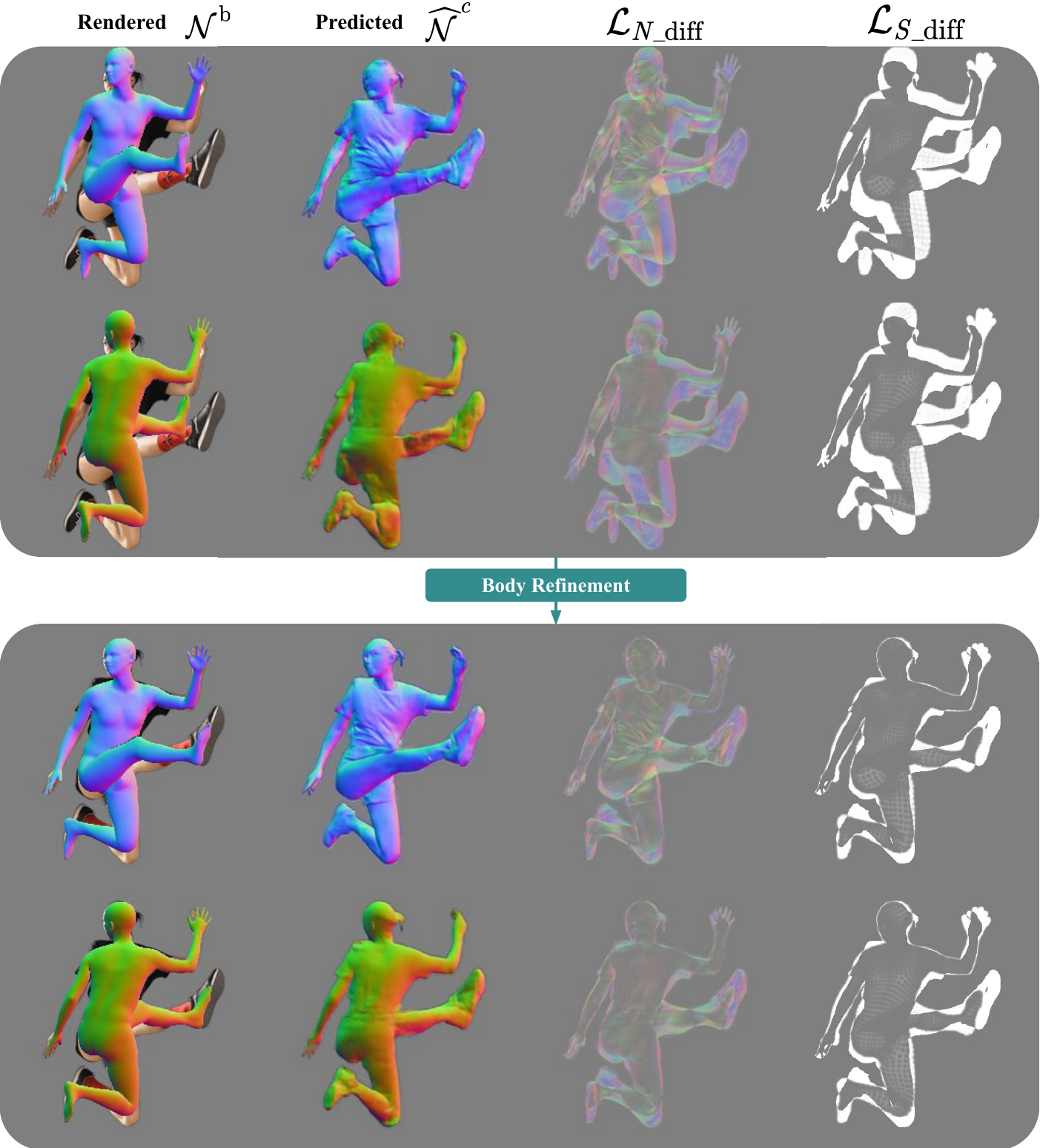}
    \vspace{-1.5em}
    \caption{\smpl refinement using a feedback loop.}
     \label{fig:normal-body-refinement}
    \vspace{-1.5 em}
\end{figure}

\medskip
\qheading{Refining \smpl.}
\label{sec: body-refinement}
Intuitively, a more accurate \smpl body fit provides a better prior that helps infer better clothed-body normals.
However, in practice, human pose and shape (HPS) regressors do not give pixel-aligned \smpl fits. 
To account for this, during inference, the \smpl fits are optimized based on the difference between the rendered \smpl-body normal maps, $\bodyNormImgFB$, and the predicted clothed-body normal maps, $\predCloNormImgFB$, as shown in \cref{fig:normal-body-refinement}. 
Specifically we optimize over \smpl's shape, $\beta$, pose, $\theta$, and translation, $t$, parameters to minimize:
\begin{align}
    \label{eq:body-fit}
    \mathcal{L}_\text{\smpl} = \min_{\theta, \beta, t}( \lambda_\text{N\_diff}\mathcal{L}_{\normal\text{\_diff}} + \mathcal{L}_{\text{S}\text{\_diff}}) \text{,}   \\
    \mathcal{L}_{\normal\text{\_diff}}  = | \bodyNormImgFB-\predCloNormImgFB | \text{,} \quad
    \mathcal{L}_{\text{S}\text{\_diff}} = | \mathcal{S}^{\body}-\widehat{\mathcal{S}}^{\cloth} | \text{,}
\end{align}
where $\mathcal{L}_{\normal\text{\_diff}}$ is a normal-map 
loss (L1), weighted by $\lambda_\text{N\_diff}$; $\mathcal{L}_{\text{S}\text{\_diff}}$ is a loss (L1) between the silhouettes of the \smpl body normal-map $\mathcal{S}^{\body}$ and 
the \highlight{human mask 
$\widehat{\mathcal{S}}^{\cloth}$ 
segmented~\cite{rembg}
from 
$\mathcal{I}$}. 
\highlight{We ablate 
$\mathcal{L}_{\normal\text{\_diff}}$, 
$\mathcal{L}_{\text{S}\text{\_diff}}$
in \suppl}

\qheading{Refining normals.} 
The normal maps rendered from the refined \smpl mesh, $\bodyNormImgFB$, are fed to the $\normNet$ networks. 
The improved \smpl-mesh-to-image alignment guides $\normNet$ to infer more reliable and detailed normals $\predCloNormImgFB$.

\qheading{Refinement loop.}
During inference, \modelname alternates between:
(1) refining the \smpl mesh using the inferred $\predCloNormImgFB$ normals and 
(2) re-inferring $\predCloNormImgFB$ using the refined \smpl. 
Experiments show that this feedback loop leads to more reliable clothed-body normal maps 
for both (front/back) sides.

\subsection{Local-feature based implicit \threeD reconstruction}
\label{sec: local-feature based implicit surface regressor}

Given the predicted clothed-body normal maps, 
$\predCloNormImgFB$, 
and the \smpl-body mesh, $\bodyMesh$, 
we regress the implicit \threeD surface of a clothed human 
based on local features $\pointFeat$:
\begin{equation}
\label{eq:point feature}
    \pointFeat = [
    \bodySDF(\text{P}), 
    \bodyNormFeat(\text{P}), 
    \cloNormFeat(\text{P})
    ],
\end{equation}
where 
$\bodySDF$ is the signed distance 
from a query point $\text{P}$ to the closest body point $\text{P}^\body \in \bodyMesh$, 
\highlight{and} 
$\bodyNormFeat$ is the barycentric surface normal of 
$\text{P}^\body$; \highlight{both provide strong regularization against self occlusions}. 
\highlight{Finally,} 
$\cloNormFeat$ is a normal vector extracted from $\predCloNormImg_\text{front}$ or $\predCloNormImg_\text{back}$ depending on the visibility of $\text{P}^\body$:
\begin{equation}
\label{eq:visibility_guided_cloNorm}
    \cloNormFeat(\text{P}) = 
    \begin{cases}
        \predCloNormImg_\text{front}(\pi(\text{P})) \text{\quad if $\text{P}^\body$ is visible}\\
        \predCloNormImg_\text{back}(\pi(\text{P}))  \text{\quad else,}
    \end{cases}
\end{equation}
where $\pi(\text{P})$ denotes the \twoD projection of the \threeD point $\text{P}$. 

Please note that $\pointFeat$ is \emph{independent} of global body pose.
Experiments show that this is key for robustness to \ood poses and efficacy \wrt training data.

We feed $\pointFeat$ into an implicit function, $\imFunc$, parameterized by a Multi-Layer Perceptron (MLP) to estimate the occupancy at point $\text{P}$, denoted as $\widehat{o}(\text{P})$. 
A mean squared error loss is used to train $\imFunc$ with \groundtruth occupancy, $o(\text{P})$. 
Then the fast surface localization algorithm~\cite{li2020realtimeVolMoCap, lorensen1987marching} is used to extract meshes from the \threeD occupancy inferred by $\imFunc$.

\section{Experiments}
\label{sec: experiments}

\subsection{Baseline models}
\label{sec: baseline models}
We compare \modelname primarily with \pifu~\cite{saito2019pifu} and \pamir~\cite{zheng2020pamir}. 
These methods differ from \modelname and from each other \wrt the training data, the loss functions, the network structure, the use of the \smpl body prior, \etc. 
To isolate and evaluate each factor, we re-implement \pifu and \pamir by ``simulating'' them based on \modelname's architecture. 
This provides a unified benchmarking framework, and enables us to easily train each baseline with the exact same data and training hyper-parameters for a fair comparison. 
Since there might be small differences \wrt the original models, we denote the ``simulated'' models \highlight{with a ``star''} as:
\begin{itemize}[noitemsep,nolistsep]
\small
\itemsep 0.5em 
    \item   \makebox[1.1cm][l]{\pifuSIM}   :\quad\makebox[3.1cm][l]{$\{f_\text{\twoD}(\mathcal{I}, \mathcal{N})\}$}                                $\rightarrow \mathcal{O}$,
    \item   \makebox[1.1cm][l]{\pamirSIM}  :\quad\makebox[3.1cm][l]{$\{f_\text{\twoD}(\mathcal{I}, \mathcal{N}), f_\text{\threeD}(\mathcal{V})\}$} $\rightarrow \mathcal{O}$,
    \item   \makebox[1.1cm][l]{\modelname}         :\quad\makebox[3.1cm][l]{$\{\mathcal{N}, \gamma(\bodyMesh)\}$}                                          $\rightarrow \mathcal{O}$,
\end{itemize}
where 
$f_\text{\twoD}$    denotes the \twoD   image encoder, 
$f_\text{\threeD}$  denotes the \threeD voxel encoder, 
$\mathcal{V}$       denotes the voxelized \smpl,
$\mathcal{O}$       denotes the entire predicted occupancy field, and 
$\gamma$            is      the mesh-based local feature extractor described in \cref{sec: local-feature based implicit surface regressor}. 
The results are summarized in \mbox{\cref{tab:benchmark}\textcolor{red}{-A}}, and discussed in \mbox{\cref{sec: evaluation}\textcolor{red}{-A}}. 
For reference, we also report the performance of the original \pifu~\cite{saito2019pifu}, \pifuhd~\cite{saito2020pifuhd}, and \pamir~\cite{zheng2020pamir}; our ``simulated'' models perform well, and even outperform the original ones. 

\subsection{Datasets}
\label{sec: dataset}

Several public or commercial \threeD clothed-human datasets are used in the literature, but each method uses different subsets and combinations of these, as shown in \cref{tab:dataset}. 

\textbf{Training data.}
To compare models fairly, we factor out differences in training data as explained in \cref{sec: baseline models}. 
Following previous work \cite{saito2019pifu, saito2020pifuhd}, we retrain all baselines on the same $450$ \renderppl scans (subset of \agora). 
Methods that require the \threeD body prior (\ie, \pamir, \modelname) use the \smplx meshes provided by \agora. 
\modelname's $\normNet$ and $\imFunc$ modules are trained on the same data.

\textbf{Testing data.}
We evaluate primarily on  \cape~\cite{ma2020cape}, which no method uses for training, to test their generelizability. 
Specifically, we divide the \cape dataset into the \capeFP and \capeNFP sets that have \highlight{``fashion'' and ``non-fashion'' poses}, respectively, to better analyze the generalization to complex body poses; for details on data splitting please see \suppl 
To evaluate performance without a domain gap between train/test data, we also test all models on \agoraFIFTY~\cite{saito2019pifu,saito2020pifuhd}, which contains $50$ samples from \agora that are different from the $450$ used for training. 

\textbf{Generating synthetic data.}
We use the OpenGL scripts of \monoport~\cite{li2020realtimeVolMoCap} to render photo-realistic images with dynamic lighting. 
We render each clothed-human \threeD scan ($\mathcal{I}$ and $\gtCloNormImgFB$) and their \smplx fits ($\bodyNormImgFB$) from multiple views by using a weak perspective camera and rotating the scan in front of it. 
In this way we generate $138,924$ samples, 
each containing a \threeD clothed-human scan, its \smplx fit, an \rgb image, camera parameters, \twoD normal maps for the scan and the \smplx mesh (from two opposite views) 
and \smplx triangle visibility information \wrt the camera.

\newcommand{\captionDATASETS}{
            Datasets for \threeD clothed humans. 
            Gray color indicates \colorbox{Gray}{datasets} used by \modelname. 
            The bottom ``number of scans'' row indicates the number of scans each method uses.
            The cell format is {\tt ~number\_of\_scans [method]}. 
            \modelname is denoted as {\tt [IC]}. 
            The symbol {\tt $\dagger$} corresponds to the ``8x'' setting in \cref{fig:datasize}.
}

\begin{table}
\centering
\resizebox{\linewidth}{!}{
\begin{tabular}{l|c|c|a|a|c|a|}
&\multicolumn{4}{c|}{Train \& Validation Sets}&\multicolumn{2}{c|}{Test Set}  \\  \hhline{~------}
& 
\rotatebox[origin=c]{0}{Renderp.}    & 
\rotatebox[origin=c]{0}{\twindom}    & 
\rotatebox[origin=c]{0}{\agora}      & 
\rotatebox[origin=c]{0}{\thuman}     & 
\rotatebox[origin=c]{0}{\buff}       & 
\rotatebox[origin=c]{0}{\cape}       \\ 
& 
\cite{renderpeople}         & 
\cite{twindom}              & 
\cite{patel2021agora}       & 
\cite{zheng2019deephuman}   & 
\cite{zhang2017buff}        &  
\cite{ma2020cape,ponsMoll2017clothCap}\\
\shline
Free \& public     &   \xmark   &   \xmark    &  \xmark    &   \cmark   &   \cmark    &    \cmark \\ 
Diverse poses      &   \xmark   &   \xmark    &  \xmark    &   \cmark   &   \xmark    &    \cmark \\ 
Diverse identities &   \cmark   &   \cmark    &  \cmark    &   \xmark   &   \xmark    &    \xmark \\ 
SMPL(-X) poses     &   \xmark   &   \xmark    &  \cmark    &   \cmark   &   \cmark    &    \cmark \\ 
High-res texture   &   \cmark   &   \cmark    &  \cmark    &   \xmark   &   \cmark    &    \cmark \\ 
\hline
\multirow{3}{*}{Number of scans} & \multicolumn{1}{l|}{450~\cite{saito2019pifu,saito2020pifuhd}} & 1000~\cite{zheng2020pamir} & ~~450~[IC] & ~~600~[$\text{IC}^{\dagger}$]  & ~~~~~5~\cite{saito2019pifu,saito2020pifuhd} & 150~[IC] \\
\qquad \quad  & \multicolumn{1}{l|}{375~\cite{huang2020arch}} & & ~3109~[$\text{IC}^{\dagger}$] & ~600~\cite{zheng2020pamir} & \multicolumn{1}{l|}{~~~26~\cite{huang2020arch}} & \\
& & & & & \multirow{1}{*}{~300~\cite{zheng2020pamir,li2020realtimeVolMoCap}} & \\
\end{tabular}
}
\vspace{-0.5 em}
\caption{\captionDATASETS}
\vspace{-0.3 em}
\label{tab:dataset}
\end{table}

\begin{table*}
\centering{
\resizebox{1.0\linewidth}{!}{
  \begin{tabular}{c|l|c|ccc|ccc|ccc|aaa}
    & Methods & \smplx & \multicolumn{3}{c|}{\agora-50} & \multicolumn{3}{c|}{\cape-FP} & \multicolumn{3}{c|}{\cape-NFP} & \multicolumn{3}{a}{\cape}\\
    & & condition. & Chamfer $\downarrow$ & P2S $\downarrow$ & Normals $\downarrow$ & Chamfer $\downarrow$ & P2S $\downarrow$ & Normals $\downarrow$ & Chamfer $\downarrow$ & P2S $\downarrow$ & Normals $\downarrow$ & Chamfer $\downarrow$ & P2S $\downarrow$ & Normals $\downarrow$\\
    \shline
    Ours & $\text{\modelname}$ & \cmark  & 1.204 & 1.584 & 0.060 & 1.233 & \textbf{1.170} & 0.072 & 1.096 & \textbf{1.013} & 0.063 & \textbf{1.142} & \textbf{1.065} & 0.066\\
    \hline
    \multirow{6}{*}{A} 
    & \makebox[1.3cm][l]{$\text{\pifu}$}~\cite{saito2019pifu} &  \xmark & 3.453 & 3.660 & 0.094 & 2.823 & 2.796 & 0.100 & 4.029 & 4.195 & 0.124 & 3.627 & 3.729 & 0.116\\
    & \makebox[1.3cm][l]{$\text{\pifuhd}$}~\cite{saito2020pifuhd} &  \xmark & 3.119 & 3.333 & 0.085 & 2.302 & 2.335 & 0.090 & 3.704 & 3.517 & 0.123 & 3.237 & 3.123 & 0.112\\
    & \makebox[1.3cm][l]{$\text{\pamir}$}~\cite{zheng2020pamir} & \cmark & 2.035 & 1.873 & 0.079 & 1.936 & 1.263 & 0.078 & 2.216 & 1.611 & 0.093 & 2.122 & 1.495 & 0.088\\
    \hhline{~--------------}
    & $\text{\smplx GT}$ &  N/A & 1.518 & 1.985 & 0.072 & 1.335 & 1.259 & 0.085 & \textbf{1.070} & 1.058 & 0.068 & 1.158 & 1.125 & 0.074\\
    \hhline{~--------------}
    & $\text{\pifuSIM}$ &  \xmark & 2.688 & 2.573 & 0.097 & 2.100 & 2.093 & 0.091 & 2.973 & 2.940 & 0.111 & 2.682 & 2.658 & 0.104\\
    & $\text{\pamirSIM}$ & \cmark & 1.401 & 1.500 & 0.063 & 1.225 & 1.206 & \textbf{0.055} & 1.413 & 1.321 & 0.063 & 1.350 & 1.283 & \textbf{0.060}\\
    \hline
    \multirow{2}{*}{B} 
    & $\text{\modelname}_{\text{N}^{\dagger}}$ & \cmark & \textbf{1.153} & 1.545 & 0.057 & 1.240 & 1.226 & 0.069 & 1.114 & 1.097 & 0.062 & 1.156 & 1.140 & 0.064\\
    & \modelname w/o $\mathcal{F}_\text{n}^\text{b}$ & \cmark & 1.259 & 1.667 & 0.062 & 1.344 & 1.336 & 0.072 & 1.180 & 1.172 & 0.064 & 1.235 & 1.227 & 0.067 \\ 
    \hline
    \multirow{2}{*}{C} 
    & $\text{\modelname}_{\text{enc}(\mathcal{I},\predCloNormImgFB)}$ & \cmark  & 1.172 & \textbf{1.350} & \textbf{0.053} & 1.243 & 1.243 & 0.062 & 1.254 & 1.122 & 0.060 & 1.250 & 1.229 & 0.061 \\
    & $\text{\modelname}_{\text{enc}(\predCloNormImgFB)}$ & \cmark  & 1.180 & 1.450 & 0.055 & \textbf{1.202} & 1.196 & 0.061 & 1.180 & 1.067 & \textbf{0.059} & 1.187 & 1.110 & \textbf{0.060}\\
    \hline
    \multirow{4}{*}{D} 
    & $\text{\modelname}$ & \wcmark & 1.583 & 1.987 & 0.079 & 1.364 & 1.403 & 0.080 & 1.444 & 1.453 & 0.083 & 1.417 & 1.436 & 0.082\\
    & $\text{\modelname} + \text{BR}$ & \wcmark & 1.554 & 1.961 & 0.074 & 1.314 & 1.356 & 0.070 & 1.351 & 1.390 & 0.073 & 1.339 & 1.378 & 0.072\\
    & $\text{\pamirSIM}$ & \wcmark & 1.674 & 1.802 & 0.075 & 1.608 & 1.625 & 0.072 & 1.803 & 1.764 & 0.079 & 1.738 & 1.718 & 0.077\\
    \hhline{~--------------}
    & $\text{\smplx perturbed}$ & N/A & 1.984 & 2.471 & 0.098 & 1.488 & 1.531 & 0.095 & 1.493 & 1.534 & 0.098 & 1.491 & 1.533 & 0.097\\
    \hline
  \end{tabular}}
}
\vspace{-0.5 em}
\caption{
    Quantitative \change{evaluation} 
    (cm) for: 
    \textbf{(A)} performance \wrt SOTA; 
    \textbf{(B)} body-guided normal prediction; 
    \textbf{(C)} local-feature based implicit reconstruction; and 
    \textbf{(D)} robustness to \smplx noise. 
    Inference conditioned on: 
    (\cmark)    \smplx ground truth (GT); 
    (\wcmark)   perturbed \smplx GT; 
    (\xmark)    no \smplx condition. 
    \smplx ground truth is provided by each dataset. 
    \colorbox{Gray}{\cape} is not used for training, and tests generalizability.
}
\label{tab:benchmark}
\end{table*}

\subsection{Evaluation}
\label{sec: evaluation}

We use 3 evaluation metrics, described in the following:
\qheading{``Chamfer'' distance.}
We report the Chamfer distance between \groundtruth scans and \highlight{reconstructed} 
meshes. 
For this, we sample points uniformly on scans/meshes, to factor out resolution differences, and compute average bi-directional point-to-surface distances. 
This metric captures large geometric differences, but misses smaller geometric details.

\qheading{``P2S'' distance.}
\cape has raw scans as ground truth, which can contain large holes. 
To factor holes out, we additionally report the average point-to-surface (P2S) distance from scan points to the closest reconstructed surface points. 
This metric can be viewed as a 
\change{1-directional} 
version of the above metric.

\qheading{``Normals'' difference.}
\change{We render normal images for reconstructed and \groundtruth surfaces from fixed viewpoints (\cref{sec: dataset}, ``generating synthetic data''), and calculate the L2 error between them. 
This captures errors for high-frequency geometric details, when Chamfer and P2S errors are small.}

\textbf{A. \modelname~-vs-~\sota.}
\modelname outperforms all original \stateoftheart (\sota) methods, and is competitive to our ``simulated'' versions of them, as shown in \mbox{\cref{tab:benchmark}\textcolor{red}{-A}}. 
We use \agora's \smplx \cite{patel2021agora} ground truth (GT) as a reference. 
We notice that our re-implemented \pamirSIM outperform the \smplx GT for images with in-distribution body poses (\agoraFIFTY and \capeFP), 
However, this is not the case for images with \ood poses (\capeNFP). 
This shows that, although conditioned on \change{GT}~\smplx fits, \pamirSIM is still sensitive to global body pose due to its global feature encoder, and fails to generalize to \ood poses. 
On the contrary, \modelname generalizes well to \ood poses, because its local features are independent from 
global pose (see \cref{sec: local-feature based implicit surface regressor}).

\begin{figure}
    \centering
    \vspace{-0.5 em}
    \includegraphics[trim=003mm 000mm 000mm 003mm, clip=true, width=1.0\linewidth]{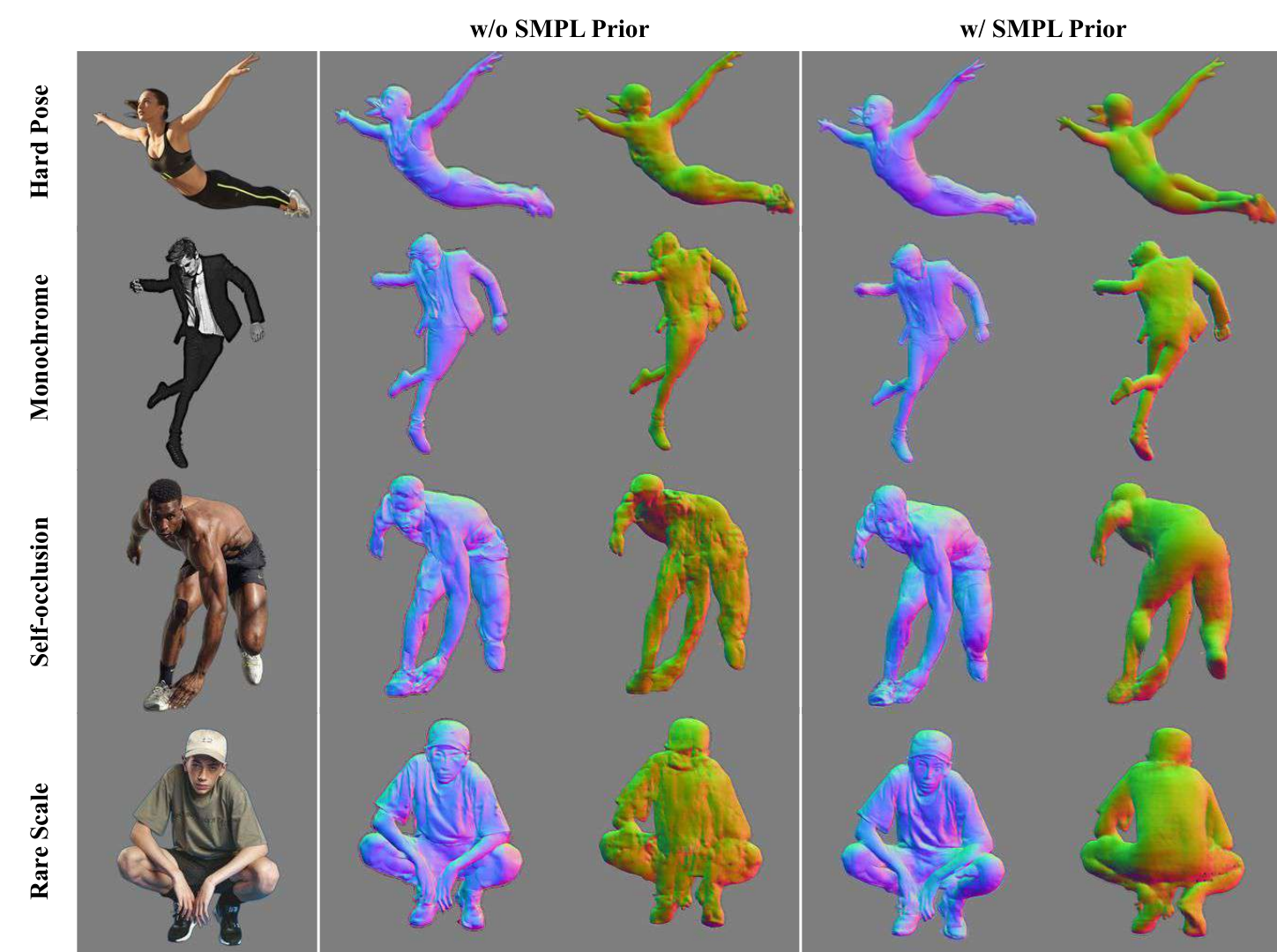}
    \vspace{-1.7 em}
    \caption{Normal prediction ($\predCloNormImg$) w/ and w/o \smpl prior ($\bodyNormImgFB$).}
    \label{fig:normal-prior}
    \vspace{-0.5 em}
\end{figure}

\textbf{B. Body-guided normal prediction.}
We evaluate the conditioning on \smplx-body normal maps, $\bodyNormImgFB$, for guiding inference of clothed-body normal maps, $\predCloNormImg$ (\cref{sec: smpl-conditioned normal predict}). 
%
\mbox{\Cref{tab:benchmark}\textcolor{red}{-B}}
\change{shows performance} with (``\modelname'') and without (``$\text{\modelname}_{\text{N}^{\dagger}}$'') conditioning. 
With no conditioning, errors on ``\cape'' increase slightly. 
Qualitatively, guidance by 
body normals 
\change{heavily improves} 
the inferred normals, especially for occluded body regions; see \cref{fig:normal-prior}.
We also \change{ablate} 
the \change{effect} 
of the body-normal feature (\cref{sec: local-feature based implicit surface regressor}), $\mathcal{F}_\text{n}^\text{b}$, by removing it; 
\change{this worsens results, see} ``\modelname w/o $\mathcal{F}_\text{n}^\text{b}$'' in \cref{tab:benchmark}\textcolor{red}{-B}. 

\vspace{-0.2mm}
\textbf{C. Local-feature based implicit reconstruction.} 
To evaluate the importance of our ``local'' features (\cref{sec: local-feature based implicit surface regressor}), $\pointFeat$, 
we replace them with ``global'' features produced by 
\twoD convolutional filters. 
These are applied on the image and the clothed-body normal maps 
(``$\text{\modelname}_{\text{enc}(\mathcal{I},\predCloNormImgFB)}$'' in \cref{tab:benchmark}\textcolor{red}{-C}),
or only on the normal maps 
(``$\text{\modelname}_{\text{enc}(            \predCloNormImgFB)}$'' in \cref{tab:benchmark}\textcolor{red}{-C}). 
We use a 2-stack hourglass model~\cite{aaron2018volreg}, whose receptive field expands to $46\%$ of the image size. 
This takes a large image area into account and produces features sensitive to global body pose. 
This worsens reconstruction performance for \ood poses, such as in \capeNFP. 
\highlight{For an evaluation of \pamir's receptive field size, see \suppl}

\begin{figure}
\centering
    \includegraphics[trim=004mm 006mm 005mm 008.7mm, clip=true, width=1.0 \linewidth]{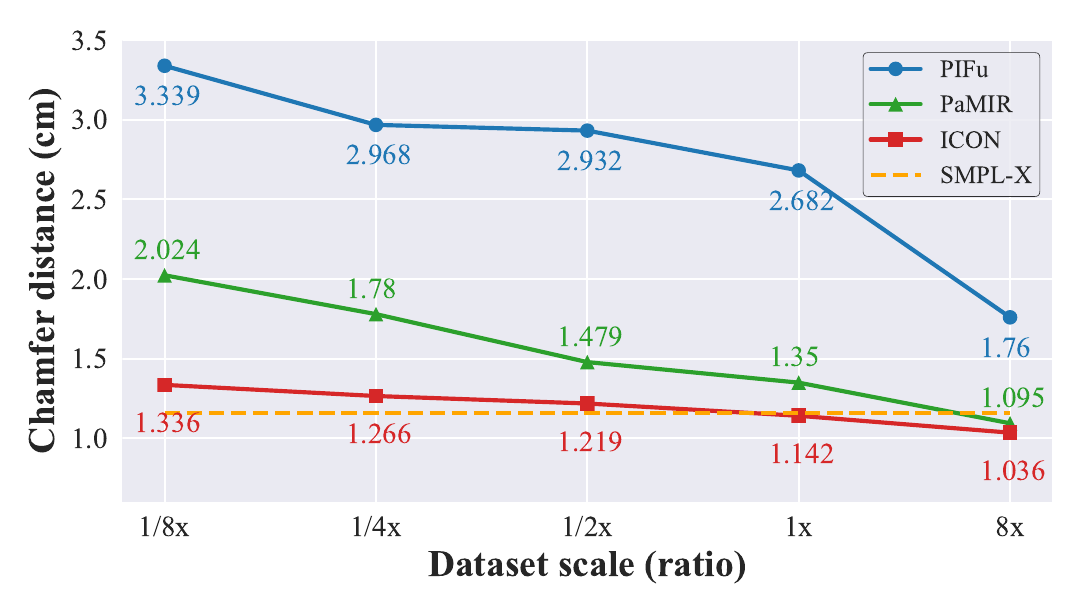}
    \caption{
                Reconstruction error \wrt training-data size. 
                ``Dataset size'' is defined as the ratio \wrt the $450$ scans used in \cite{saito2019pifu, saito2020pifuhd}. 
                The ``8x'' setting is all $3,709$ scans of \agora~\cite{patel2021agora} and \thuman~\cite{zheng2019deephuman}. 
    }
    \label{fig:datasize}
\end{figure}

We compare \modelname to \stateoftheart (\sota) models for a varying amount of training data in \cref{fig:datasize}. 
The ``Dataset scale'' axis reports the data size as the ratio 
\wrt the $450$ scans of the original \pifu methods \cite{saito2019pifu, saito2020pifuhd}; the left-most side corresponds to $56$ scans and the right-most side corresponds to $3,709$ scans, \ie, all the scans of \agora~\cite{patel2021agora} and \thuman~\cite{zheng2019deephuman}. 
\modelname consistently outperforms all methods. 
Importantly, \modelname achieves \sota performance even when trained on just a \emph{fraction} of the data. 
We attribute this to the local nature of \modelname's point features; this helps \modelname generalize well in the pose space and be data efficient.

\textbf{D. Robustness to \smplx noise.}
\smplx estimated from an image might not be perfectly aligned with body pixels \change{in the image}. 
\change{However, \pamir and \modelname are conditioned on this estimation.}
Thus, 
\change{they}
need to be robust against \highlight{various noise levels} in \smplx shape and pose. 
To evaluate this, 
we feed \pamirSIM and \modelname with \groundtruth and perturbed \smplx, denoted with (\cmark) and (\wcmark) in \mbox{\cref{tab:benchmark}\textcolor{red}{-A,D}}.
\modelname conditioned on perturbed (\wcmark) \smplx produces larger errors \wrt conditioning on ground truth (\cmark). 
However, adding the body refinement module (``\modelname+BR'') of \cref{sec: smpl-conditioned normal predict}, refines \smplx and improves performance.
As a result, ``\modelname+BR'' conditioned on noisy \smplx (\wcmark) performs comparably to \pamirSIM conditioned on \groundtruth \smplx (\cmark); 
it is slightly worse/better for in-/\ood poses.

\section{Applications}
\label{sec: applications}

\subsection{Reconstruction from in-the-wild images}
\label{sec: application ICON}

We collect $200$ in-the-wild images from Pinterest that show people performing parkour, sports, street dance, and kung fu. 
These images are unseen during training. 
We show qualitative results for \modelname in \cref{fig:demo} and comparisons to SOTA in \cref{fig:motivation}; for more results see our \video and \suppl 

To evaluate the perceived realism of our results, we compare \modelname to \pifuSIM, \pamirSIM, and the original \pifuhd~\cite{saito2020pifuhd} in a perceptual study. 
\modelname, \pifuSIM and \text{\pamirSIM} are trained on all $3,709$ scans of \agora~\cite{patel2021agora} and \thuman~\cite{zheng2019deephuman} (``8x'' setting in \cref{fig:datasize}). 
For \pifuhd we use its pre-trained model. 
In the study, participants were shown an image and either a rendered result of \modelname or of another method.
Participants were asked to choose the result that best represents the shape \change{of the} human in the image. 
We report the percentage of trails in which participants preferred the baseline methods over \modelname in \cref{tab: perceptual study}; \mbox{p-values} correspond to the null-hypothesis that two methods perform equally well. 
\change{For details on the study, example stimuli, catch trials, etc.~see \suppl}

\subsection{Animatable avatar creation from video}
\label{sec: application SCANimate}

\begin{figure*}
\centering{
    \begin{subfigure}{\linewidth}
        \includegraphics[trim=000mm 000mm 000mm 000mm, clip=true, width=1.0\linewidth]{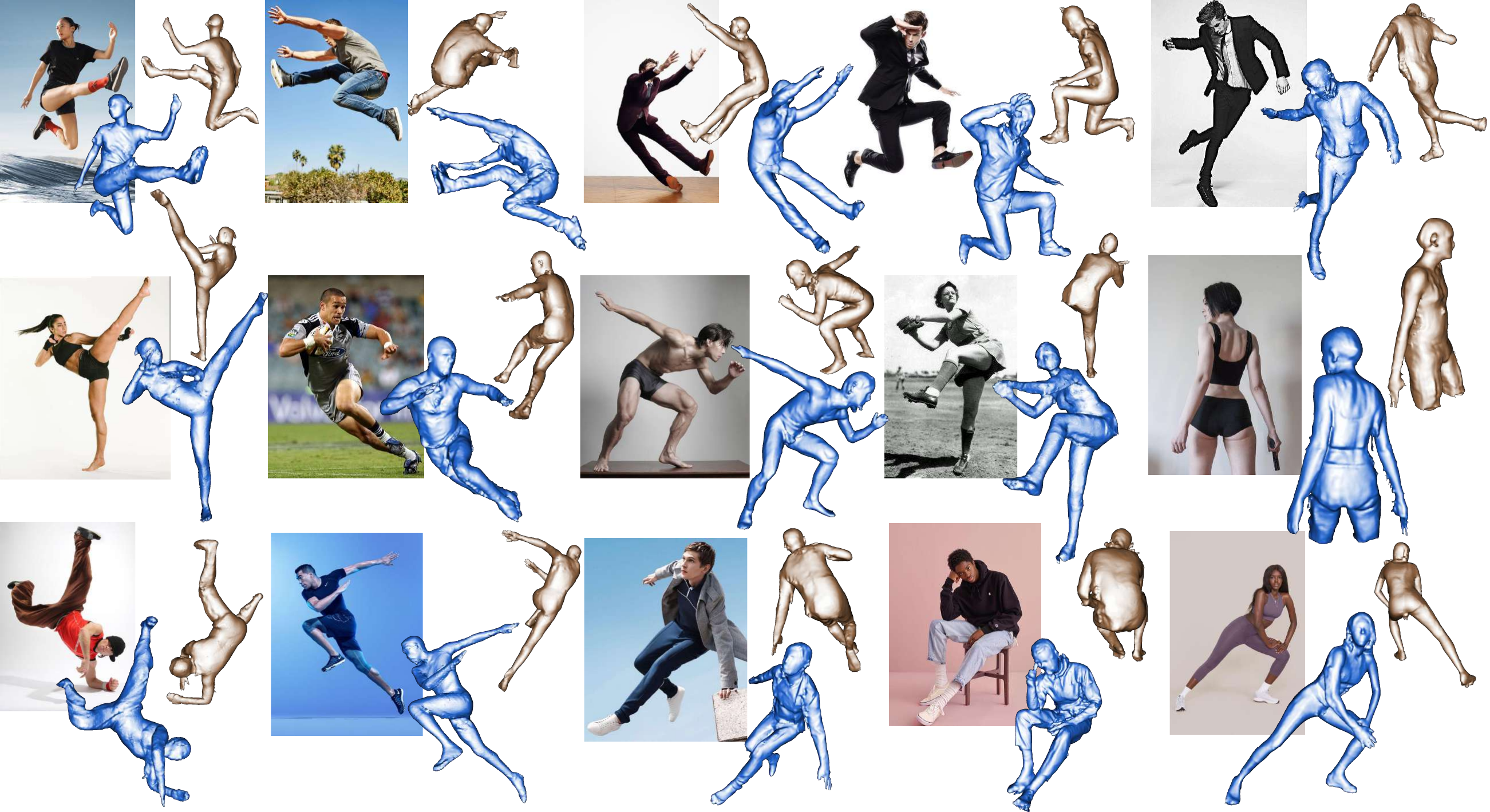}
        \caption{
                    \modelname reconstructions for in-the-wild images with extreme poses (\cref{sec: application ICON}).
        }
        \label{fig:demo}
    \end{subfigure}
    \begin{subfigure}{\linewidth}
        \includegraphics[width=\linewidth]{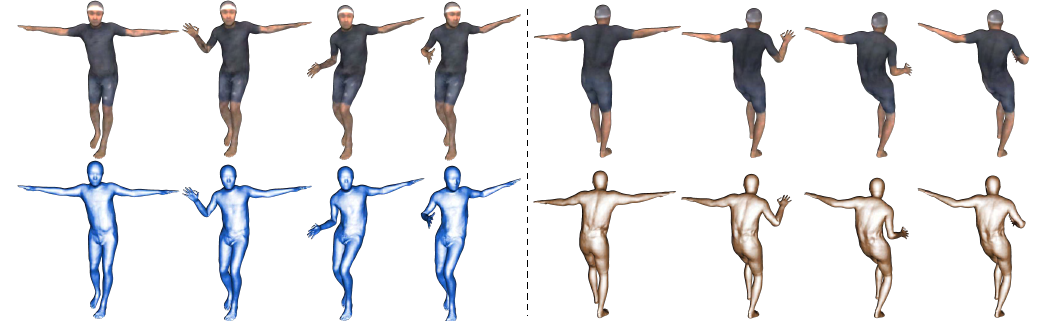}
	    \caption{
	                Avatar creation from images with \scanimate (\cref{sec: application SCANimate}).  
	                The input per-frame meshes are reconstructed with \modelname.
	    }
	    \label{fig:scanimate}
    \end{subfigure}}
    \caption{
                \modelname results for two applications (\cref{sec: applications}). 
                We show two views for each mesh, \ie, a front (\textcolor{frontcolor}{blue}) and a rotated (\textcolor{sidecolor}{bronze}) view.
    }
\end{figure*}

Given a sequence of images with the same subject in various poses, we create an animatable avatar with the help of \scanimate~\cite{saito2021scanimate}.
First, we use \modelname to reconstruct a \threeD clothed-human mesh per frame. 
Then,  we feed these meshes to \scanimate. 
\modelname's robustness to diverse poses enables us to learn a clothed avatar with pose-dependent clothing deformation.
Unlike raw \threeD scans, which are taken with multi-view systems, \modelname operates on a single image and its reconstructions are more reliable for observed body regions than for occluded ones. 
Thus, we reformulate the loss of \scanimate to downweight occluded regions \highlight{depending on camera viewpoint}. Results are shown in \cref{fig:teaser} and \cref{fig:scanimate}; for animations see the \video on our \href{https://icon.is.tue.mpg.de}{webpage}. 

\begin{figure}
\centerline{ 
\includegraphics[width=\linewidth]{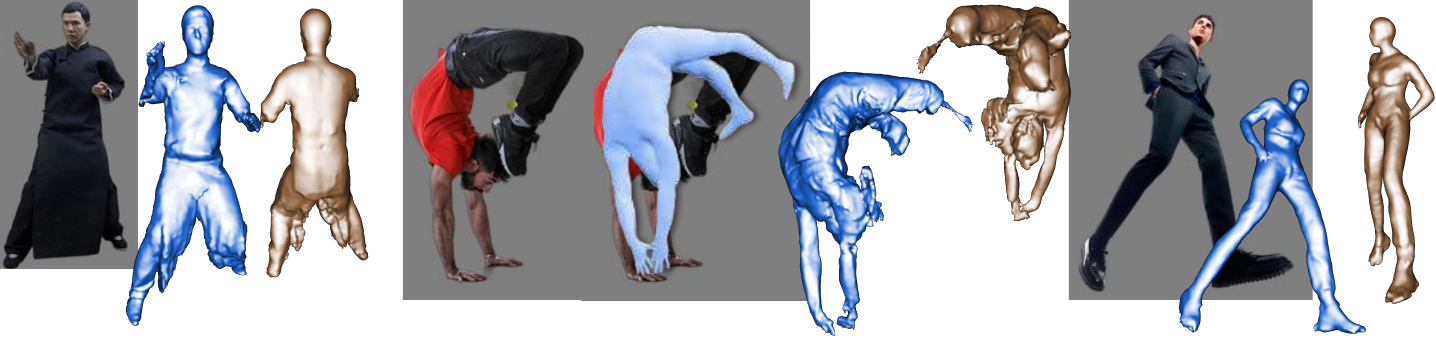}}
    \vspace{-0.5 em}
    \caption{
                Failure cases of \modelname for extreme clothing, pose, or camera view.
                We show the front (\textcolor{frontcolor}{blue}) and rotated (\textcolor{sidecolor}{bronze}) views. 
    }
    \label{fig:failure small}
\end{figure}

\begin{table}
\centering
\small
\begin{tabular}{c|ccc}
 & \pifuSIM & \pifuhd~\cite{saito2020pifuhd} & \pamirSIM \\
\shline
Preference & 30.9\% & 22.3\% & 26.6\% \\
P-value & 1.35e-33 & 1.08e-48 & 3.60e-54 \\
\end{tabular}
\caption{
            Perceptual study. 
            Numbers denote the chance that participants prefer the reconstruction of a  competing method over \modelname for \inthewild images. 
            \modelname is judged significantly more realistic. 
}
\label{tab: perceptual study}
\end{table}

\vfill

\section{Conclusion}
\label{sec:conclusion}

We have presented \modelname, which robustly recovers a \threeD clothed person from a single image with accuracy and realism that exceeds prior art. 
There are two keys: 
(1) Regularizing the solution with a 3D body model while optimizing that body model iteratively. 
(2) Using local features to eliminate spurious correlations with global pose.
Thorough ablation studies validate these choices.
The quality of results \change{is} sufficient to build a \threeD avatar from monocular image sequences.

\smallskip
\qheading{Limitations and future work.}
Due to the strong body prior exploited by \modelname, loose clothing that is far from the body may be difficult to reconstruct; see \cref{fig:failure small}.
Although \modelname is robust to small errors of body fits, significant failure of body fits leads to reconstruction failure.
Because it is trained on orthographic views, \modelname has trouble with strong perspective effects, producing asymmetric limbs or anatomically improbable shapes.
A key future application is to use images alone to create a dataset of clothed avatars.
Such a dataset could  advance research  in human shape generation~\cite{chen2022gdna}, be valuable to fashion industry, and facilitate graphics applications.

\newpage


\qheading{Possible negative impact.}
While the quality of virtual humans created from images is not at the level of facial ``deep fakes'', as this technology matures, it will open up the possibility for full-body deep fakes, with all the attendant risks.
These  risks must also be balanced by the positive use cases in entertainment, tele-presence, and future metaverse applications.
Clearly regulation will be needed to establish legal boundaries for its use.
In lieu of societal guidelines today, we have made our \href{https://github.com/YuliangXiu/ICON}{code} available with an appropriate license.

\qheading{Disclosure.}
{\small
\href{https://files.is.tue.mpg.de/black/CoI_CVPR_2022.txt}{
      https://files.is.tue.mpg.de/black/{CoI\_CVPR\_2022.txt}}}

\newpage


{
\qheading{Acknowledgments.} 
We thank Yao Feng, Soubhik Sanyal, Hongwei Yi, Qianli Ma, Chun-Hao Paul Huang, Weiyang Liu, and Xu Chen for their feedback and discussions, Tsvetelina Alexiadis for her help with perceptual study, Taylor McConnell assistance, Benjamin Pellkofer for webpage, and Yuanlu Xu's help in comparing with \arch and \archplus. 
This project has received funding from the European Union’s Horizon $2020$ research and innovation programme under the Marie Skłodowska-Curie grant agreement No.$860768$ (\href{https://www.clipe-itn.eu}{CLIPE} project).}





\clearpage

\begin{appendices}

\label{appendices}

We provide more details for the method and experiments, as well as more quantitative and qualitative results, as an extension of \cref{sec: method}, \cref{sec: experiments} and \cref{sec: applications} of the main paper.

\section{Method \& Experiment Details}
\subsection{Dataset (\cref{sec: dataset})}

\qheading{Dataset size.}
We evaluate the performance of \modelname and \sota methods for a varying training-dataset size (\cref{fig:datasize,tab:geo-datasize}).
For this, we first combine \agora~ \cite{patel2021agora} ($3,109$ scans) and \thuman~\cite{zheng2019deephuman} ($600$ scans) to get $3,709$ scans in total. 
This new dataset is $8$x times larger than the $450$ \renderppl (``$450$-Rp'') scans used in \cite{saito2019pifu,saito2020pifuhd}. 
Then, we sample this ``$8$x dataset'' to create smaller variations, for $1/8$x, $1/4$x, $1/2$x, $1$x, and $8$x the size of ``$450$-Rp''. 

\begin{figure*}[t]
  \centering
  \vspace{-3.5 em}
  \begin{subfigure}{0.32\linewidth}
    \includegraphics[width=\linewidth,frame]{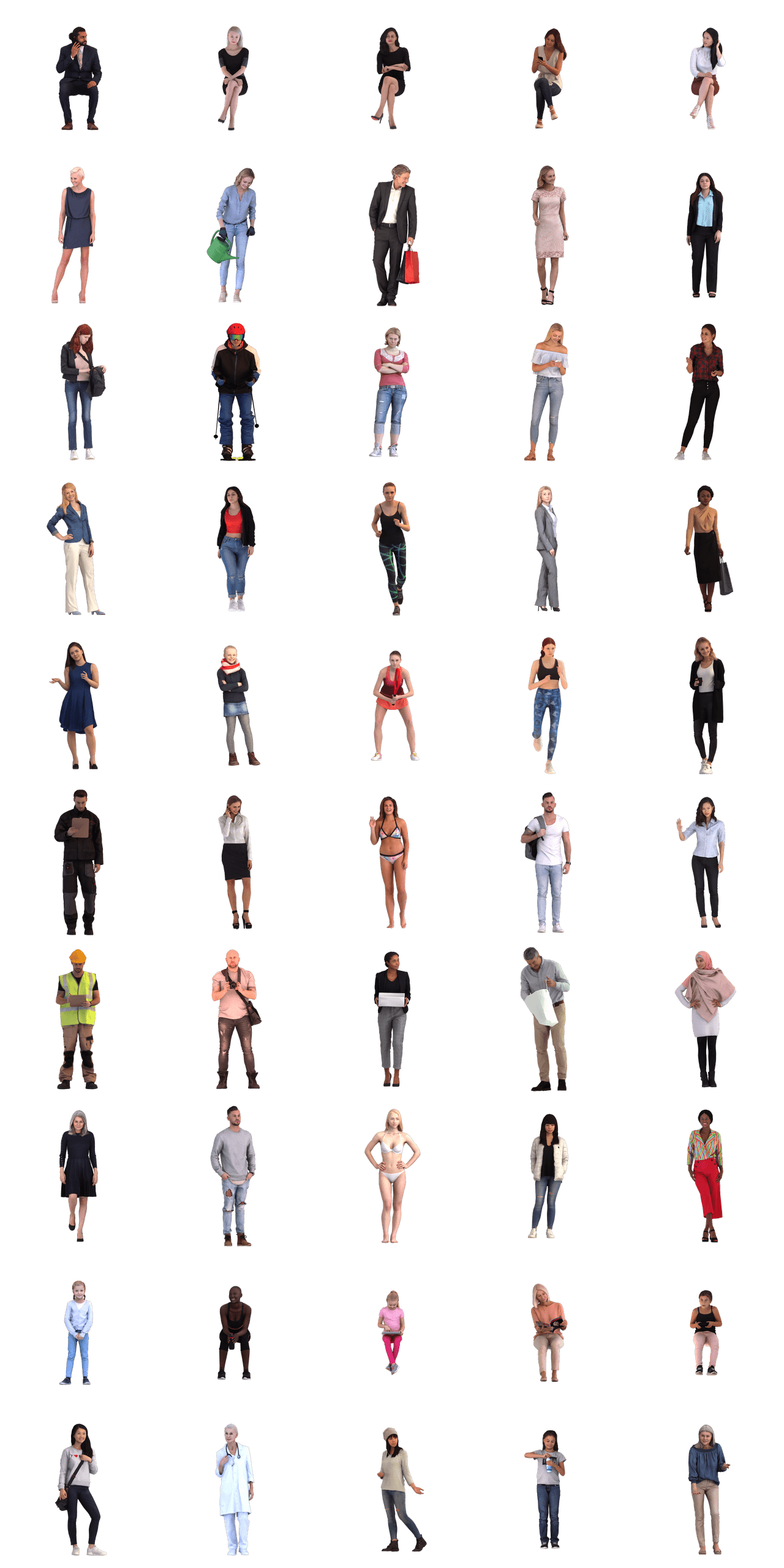}
    \caption{\renderppl~\cite{renderpeople} (450 scans)}
    \label{fig:kmeans-renderppl}
  \end{subfigure}
  \begin{subfigure}{0.64\linewidth}
    \includegraphics[width=\linewidth,frame]{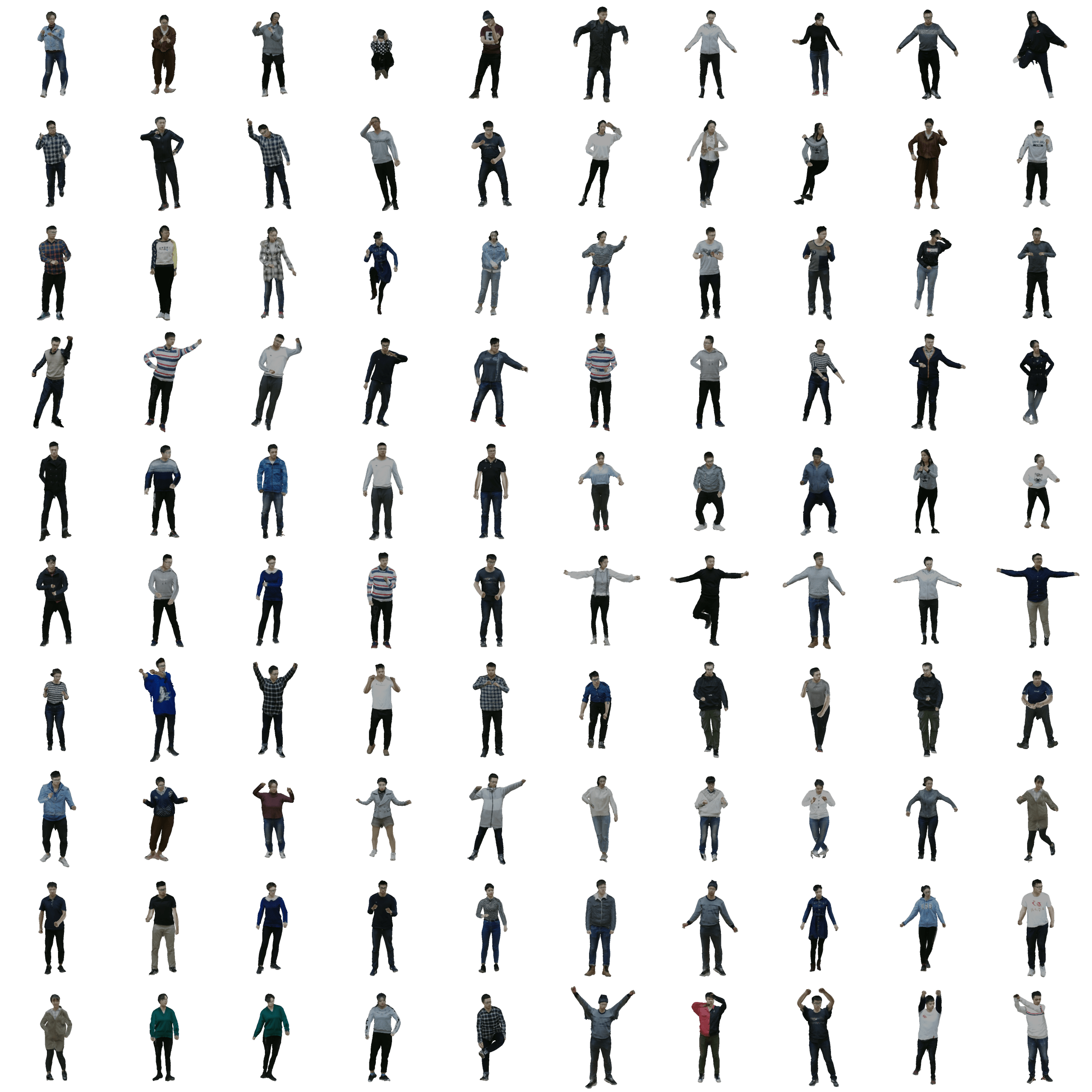}    
    \caption{\thuman~\cite{zheng2019deephuman} (600 scans)}
    \label{fig:kmeans-thuman}
  \end{subfigure}
  \begin{subfigure}{0.32\linewidth}
    \includegraphics[width=\linewidth,frame]{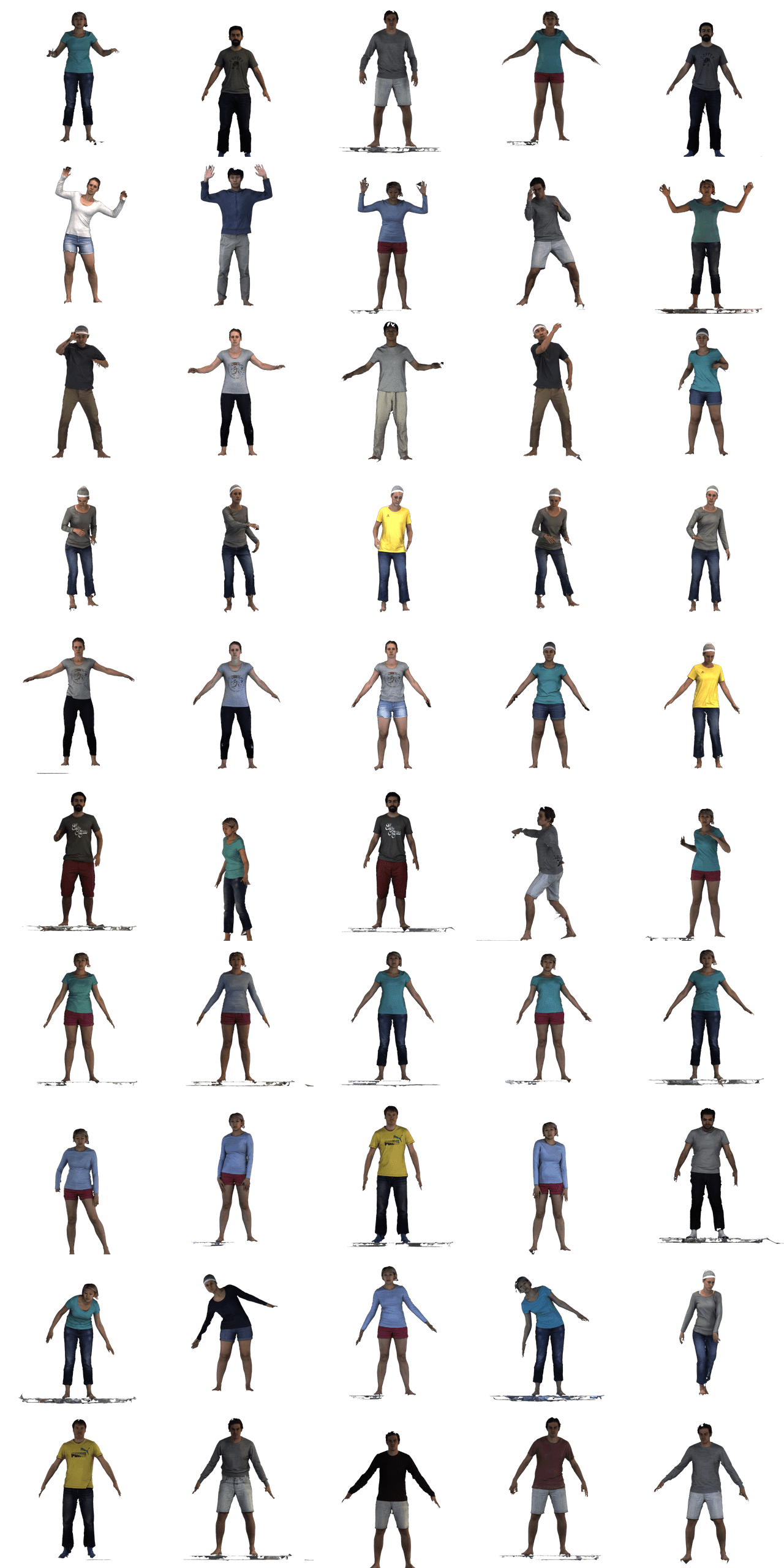}
    \caption{\capeFP~\cite{ma2020cape} (fashion poses, 50 scans)}
    \label{fig:kmeans-cape-easy}
  \end{subfigure}
  \begin{subfigure}{0.64\linewidth}
    \includegraphics[width=\linewidth,frame]{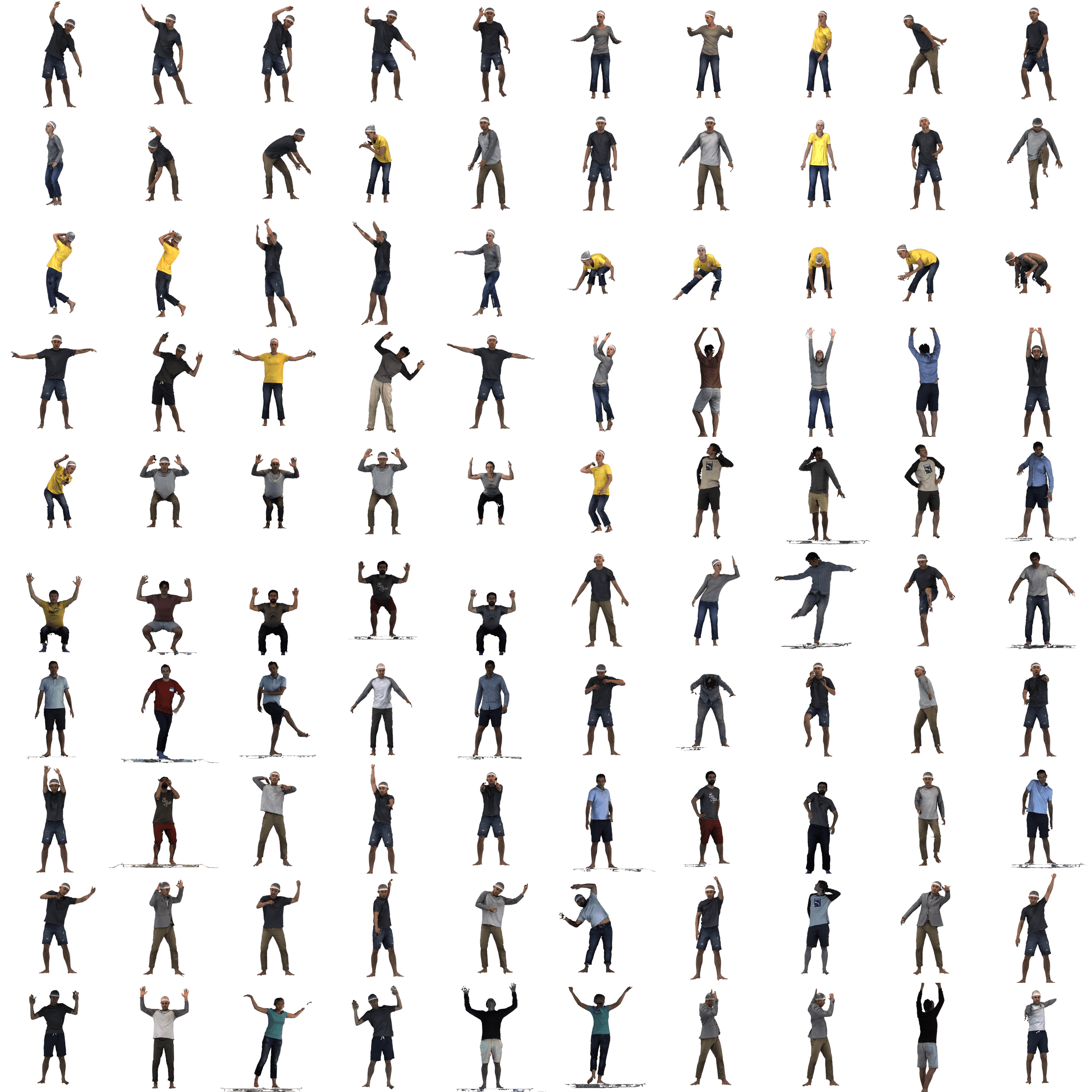}
    \caption{\capeNFP~\cite{ma2020cape} (non fashion poses, 100 scans)}
    \label{fig:kmeans-cape-hard}
  \end{subfigure}
  \caption{Representative poses for different datasets.}
  \label{fig:kmeans}
\end{figure*}

\smallskip
\qheading{Dataset splits.}
For the ``$8$x dataset'', we split the $3,109$ \agora scans into a new training set ($3,034$ scans), validation set ($25$ scans) and test set ($50$ scans). 
Among these, $1,847$ come from \renderppl~\cite{renderpeople} (see \cref{fig:kmeans-renderppl}), $622$ from \axyz~\cite{axyz}, $242$ from \humanalloy~\cite{humanalloy}, $398$ from \threeppl~\cite{3dpeople}, and we sample only $600$ scans from \thuman (see \cref{fig:kmeans-thuman}), due to its high pose repeatability and limited identity variants (see \cref{tab:dataset}), with the ``select-cluster'' scheme described below. 
These scans, as well as their \smplx fits, are rendered after every $10$ degrees rotation around the yaw axis, to totally generate {\tt $(3109 \text{~\agora} + 600 \text{~\thuman} + 150 \text{~\cape}) \times 36 = 138,924$} samples. 

\smallskip
\qheading{Dataset distribution via ``select-cluster'' scheme.} 
To create a training set with a rich pose distribution, we need to select scans from various datasets with poses different from \agora.
Following \smplify \cite{federica2016smplify}, we first fit a Gaussian Mixture Model (GMM) with $8$ components to all \agora poses, and \textbf{select} 2K \thuman scans with low likelihood. 
Then, we apply \mbox{M-Medoids} ({\tt $\text{n\_cluster} = 50$}) on these selections for \textbf{clustering}, and randomly pick $12$ scans per cluster, collecting $50 \times 12 = 600$ \thuman scans in total; see \cref{fig:kmeans-thuman}. 
This is also used to split \cape into \capeFP (\cref{fig:kmeans-cape-easy}) and \capeNFP (\cref{fig:kmeans-cape-hard}), corresponding to scans with poses similar (in-distribution poses) and dissimilar (\ood poses) to \agora ones, respectively.

\smallskip
\qheading{Perturbed \smpl.}
To perturb \smpl's pose and shape parameters, random noise is added to $\theta \text{ and } \beta$ by:
\begin{equation}
    \label{eq: perturb smpl}
    \begin{gathered}
        \theta \mathrel{+}= s_\theta * \mu  \text{,}    \\
        \beta  \mathrel{+}= s_\beta  * \mu  \text{,}    \\
    \end{gathered}
\end{equation}
where $\mu \in [-1,1]$, $s_\theta=0.15$ and $s_\beta=0.5$. 
These are set empirically to mimic the misalignment error typically caused by \offtheshell HPS during testing.

\highlight{
\smallskip
\qheading{Discussion on simulated data.}
The wide and loose clothing in \clothplus~\cite{bertiche2020cloth3d,madadi2021cloth3d++} demonstrates strong dynamics, which would complement commonly used datasets of commercial scans.
Yet, the domain gap between \clothplus and real images is still large. 
Moreover, it is unclear how to train an implicit function from multi-layer non-watertight meshes. Consequently, we leave it for future research.
}

\subsection{Refining \smpl (\cref{sec: body-refinement})}

\begin{figure}[h]
\centering
    \includegraphics[trim=006mm 002mm 006mm 002mm, clip=true, width=1.0 \linewidth]{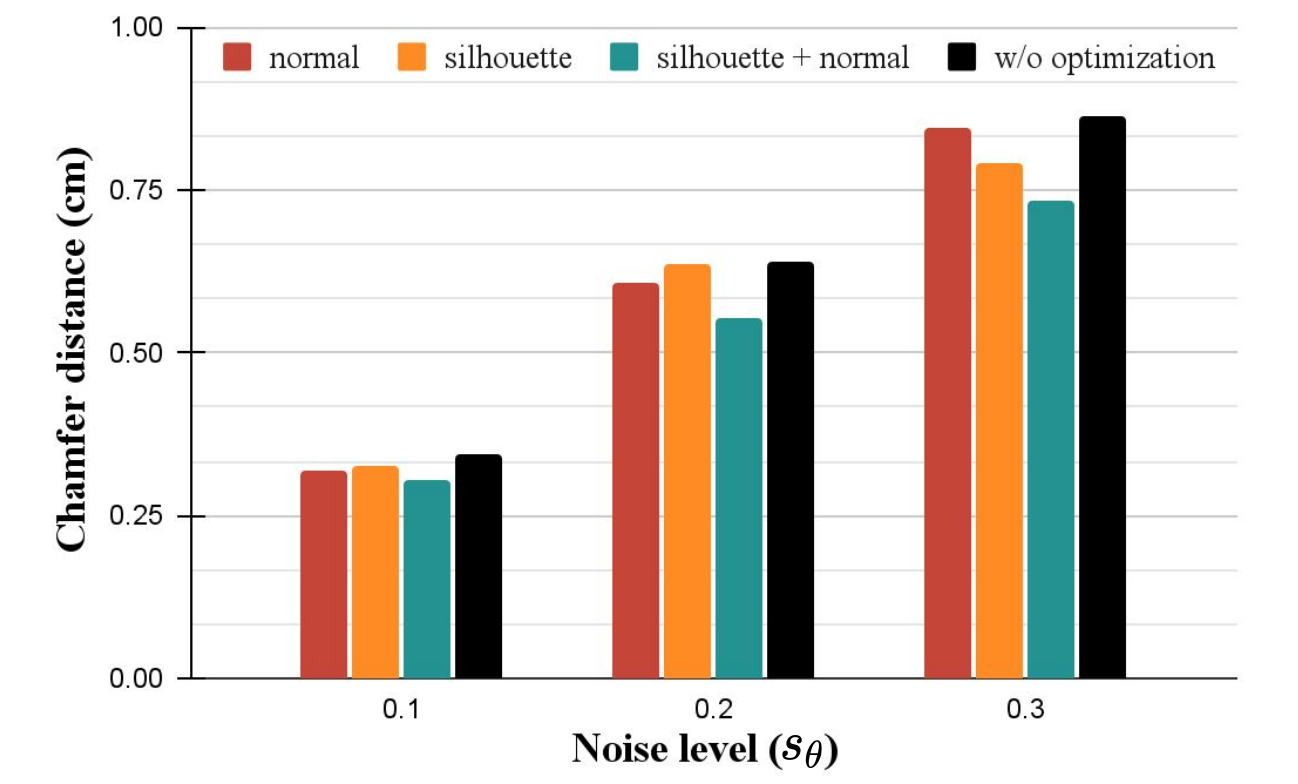}
    \caption{\smpl refinement error (y-axis) with different losses (see colors) and noise levels, $s_\theta$, of pose parameters (x-axis).}
    \label{fig:joint-optim-sanity}
\end{figure}

To statistically analyze the necessity of $\mathcal{L}_\text{N\_diff}$ and $\mathcal{L}_\text{S\_diff}$ in \cref{eq:body-fit}, we do a sanity check on \agora's validation set. 
Initialized with different pose noise, $s_\theta$ (\cref{eq: perturb smpl}), we optimize the $\{\theta, \beta, t\}$ parameters of the perturbed \smpl by minimizing the difference between rendered \smpl-body normal maps and \groundtruth clothed-body normal maps for 2K iterations. 
As \cref{fig:joint-optim-sanity} shows,  \textcolor{GreenColor}{$\mathcal{L}_\text{N\_diff} + \mathcal{L}_\text{S\_diff}$} always leads to the smallest error under any noise level, measured by the Chamfer distance between the optimized perturbed \smpl mesh and the \groundtruth \smpl mesh.

\begin{table*}
\centering
\resizebox{1.0\linewidth}{!}{
  \begin{tabular}{c|l|c|ccc|ccc|ccc|aaa}
    & Methods & \smplx & \multicolumn{3}{c|}{\agora-50} & \multicolumn{3}{c|}{\cape-FP} & \multicolumn{3}{c|}{\cape-NFP} & \multicolumn{3}{a}{\cape}\\
    & & condition. & Chamfer $\downarrow$ & P2S $\downarrow$ & Normal $\downarrow$ & Chamfer $\downarrow$ & P2S $\downarrow$ & Normal $\downarrow$ & Chamfer $\downarrow$ & P2S $\downarrow$ & Normal $\downarrow$ & Chamfer $\downarrow$ & P2S $\downarrow$ & Normal $\downarrow$\\
    \shline
     & $\text{\modelname}$ & \wcmark & 1.583 & 1.987 & 0.079 & \textbf{1.364} & \textbf{1.403} & 0.080 & \textbf{1.444} & \textbf{1.453} & 0.083 & \textbf{1.417} & \textbf{1.436} & 0.082\\
    \hline
    \multirow{4}{*}{D} & $\text{\smplx perturbed}$ & \wcmark & 1.984 & 2.471 & 0.098 & 1.488 & 1.531 & 0.095 & 1.493 & 1.534 & 0.098 & 1.491 & 1.533 & 0.097\\
    & $\text{\modelname}_\text{enc(I,N)}$ & \wcmark & 1.569 & \textbf{1.784} & \textbf{0.073} & 1.379 & 1.498 & \textbf{0.070} & 1.600 & 1.580 & \textbf{0.078} & 1.526 & 1.553 & \textbf{0.075}\\
    & $\text{\modelname}_\text{enc(N)}$ & \wcmark & \textbf{1.564} & 1.854 & 0.074 & 1.368 & 1.484 & 0.071 & 1.526 & 1.524 & \textbf{0.078} & 1.473 & 1.511 & 0.076\\
    & $\text{\modelname}_{\text{N}^{\dagger}}$ & \wcmark & 1.575 & 2.016 & 0.077 & 1.376 & 1.496 & 0.076 & 1.458 & 1.569 & 0.080 & 1.431 & 1.545 & 0.079
  \end{tabular}
}
\vspace{-0.5 em}
\caption{Quantitative errors (cm) for several \modelname variants conditioned on perturbed \smplx fits ($s_\theta = 0.15, s_\beta = 0.5$).}
\label{tab:sup-benchmark}
\end{table*}

\subsection{Perceptual study (\cref{tab: perceptual study})}

\begin{figure*}
     \centering
    \begin{subfigure}{0.49\linewidth}
        \includegraphics[width=\linewidth]{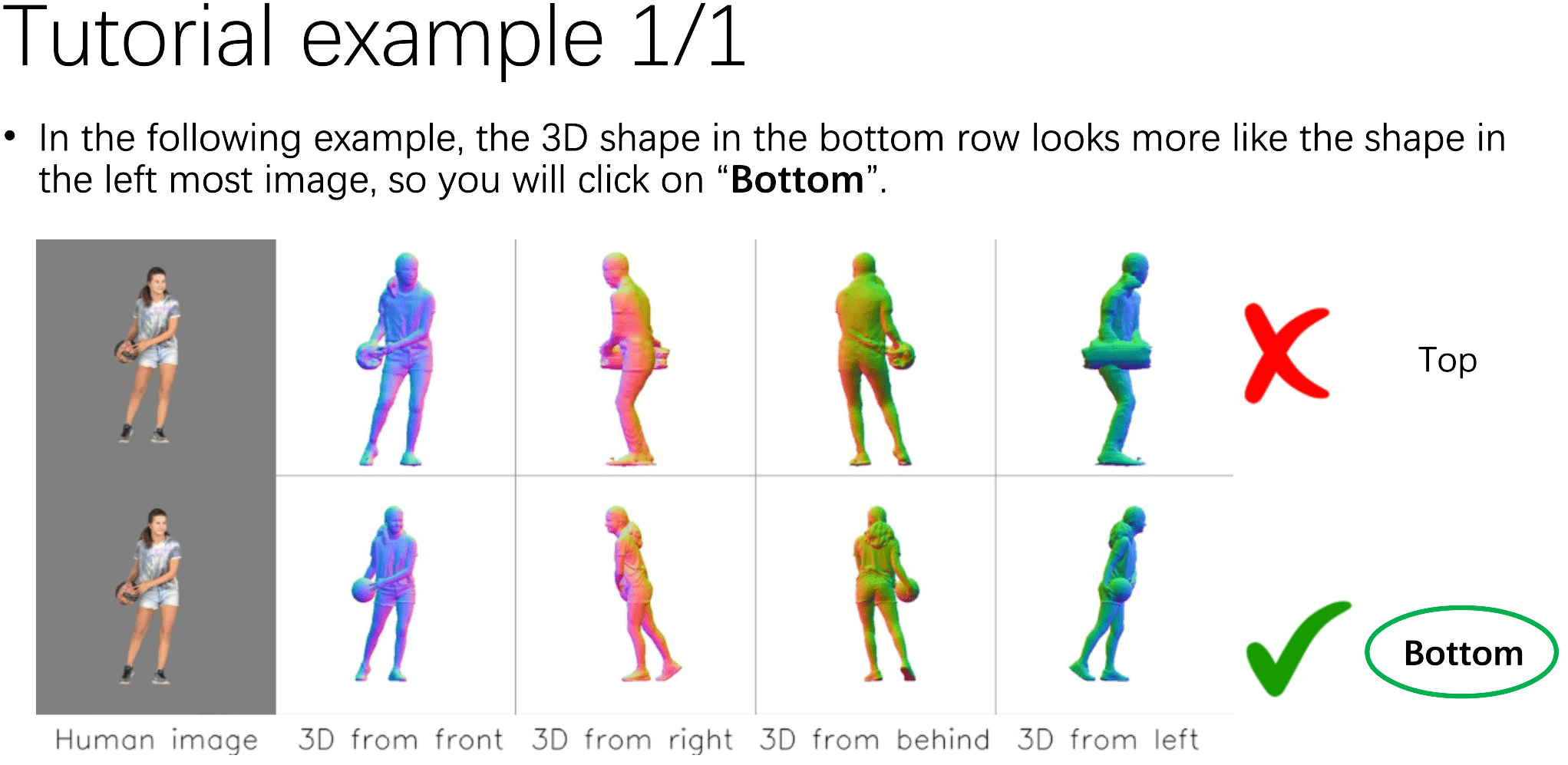}
        \caption{A tutorial sample.}
    \label{fig:perceptual_study:tutorial}
    \end{subfigure}
    \hfill
    \begin{subfigure}{0.49\linewidth}
        \includegraphics[width=\linewidth]{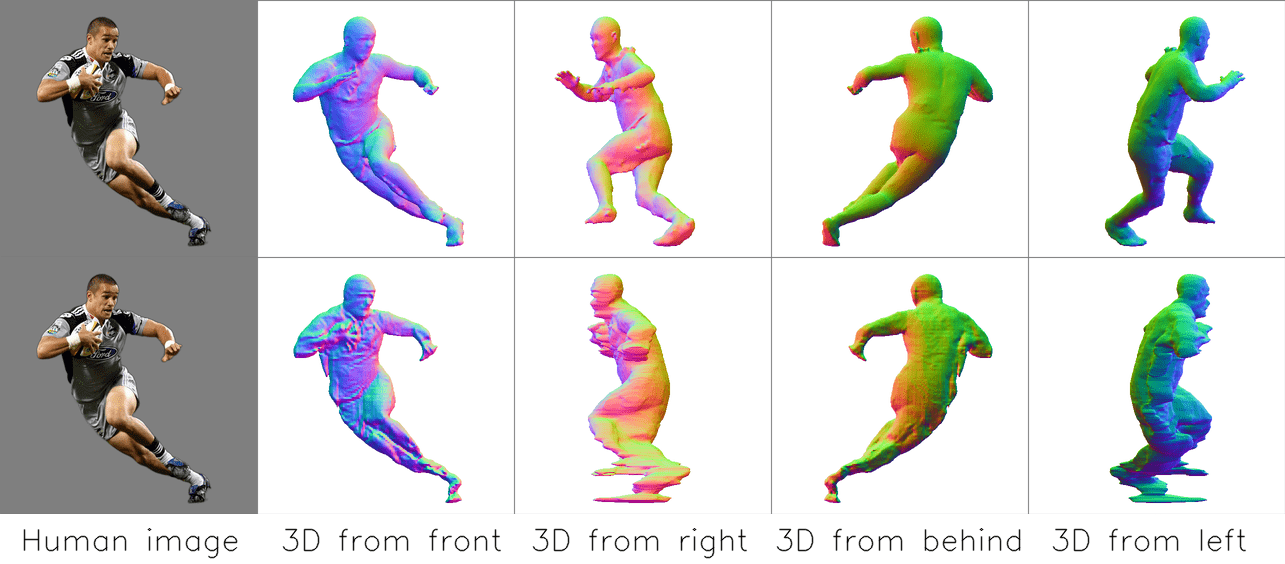}    
        \caption{An evaluation sample.}
    \label{fig:perceptual_study:evaluation}
    \end{subfigure}
    \centering
    \begin{subfigure}{\linewidth}
        \includegraphics[width=0.48\linewidth]{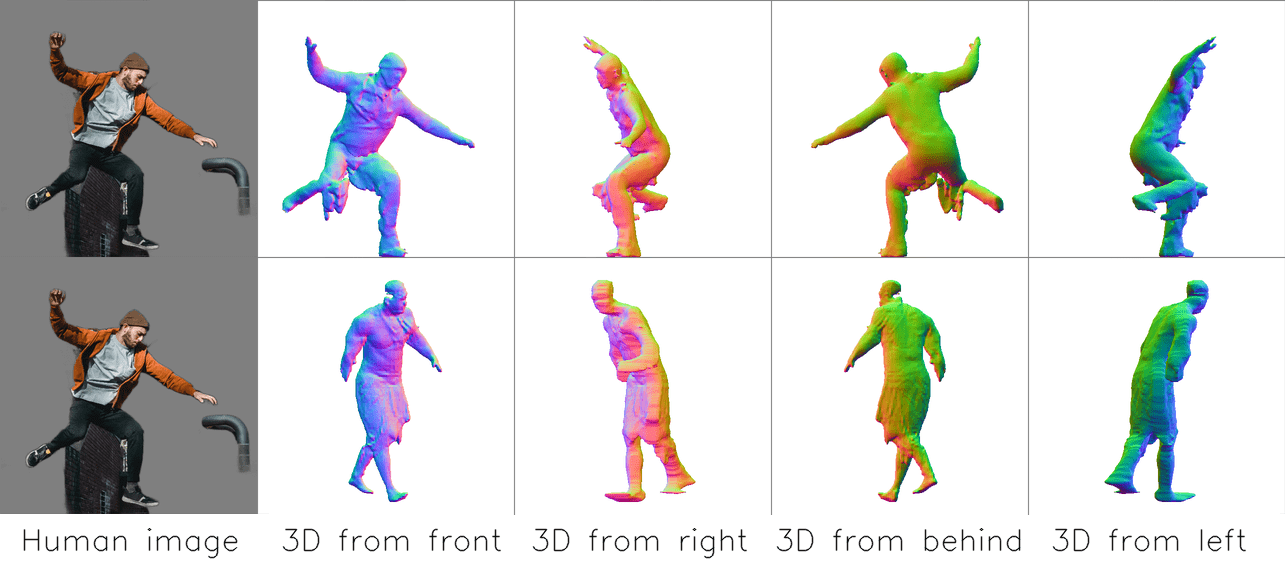}
        \hfill
        \includegraphics[width=0.48\linewidth]{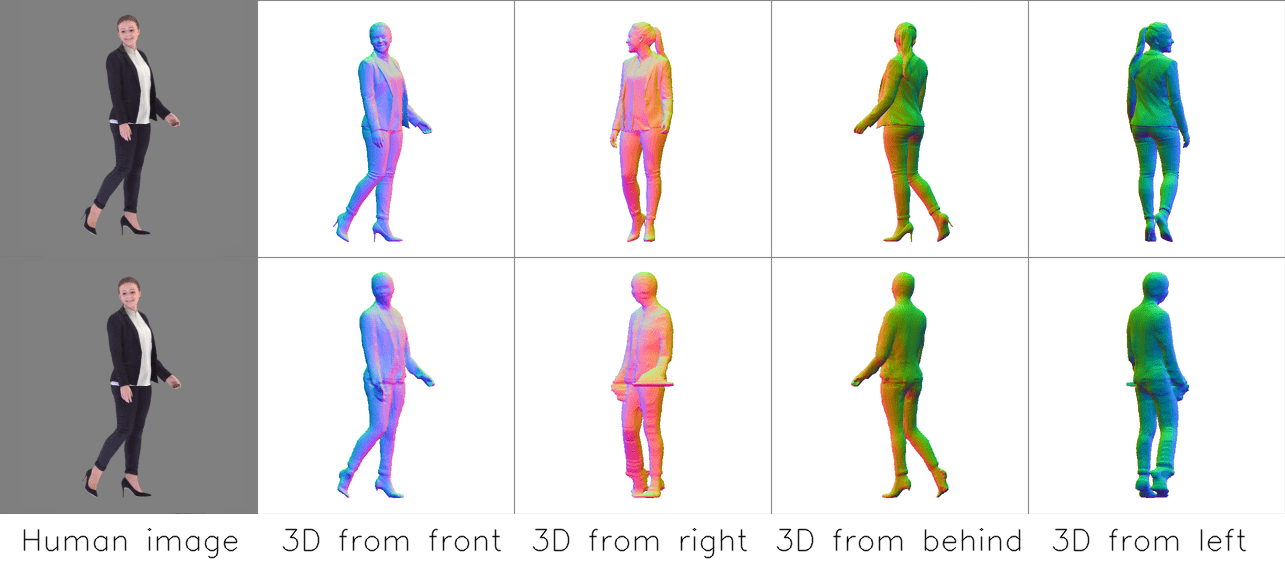} 
        \caption{   Two samples of catch trials. 
                    Left:  result from this image (top) vs from another image (bottom). 
                    Right: \groundtruth (top) vs reconstruction mesh (bottom).}
        \label{fig:perceptual_study:catch_trial}
    \end{subfigure}
    
    \caption{Some samples in the perceptual study to evaluate \textbf{reconstructions} on \inthewild images.}
    \label{fig:perceptual_study}
\end{figure*}

\qheading{Reconstruction on \inthewild images.}
We perform a perceptual study to evaluate the perceived realism of the reconstructed clothed \threeD humans from \inthewild images. 
\modelname is compared against $3$ methods, \pifu~\cite{saito2019pifu}, \pifuhd~\cite{saito2020pifuhd}, and \pamir~\cite{zheng2020pamir}.
We create a benchmark of $200$ unseen images downloaded from the internet, and apply all the methods on this test set.
All the reconstruction results are evaluated on \ac{amt}, where each participant is shown pairs of reconstructions from \modelname and one of the baselines, see \cref{fig:perceptual_study}. 
Each reconstruction result is rendered in four views: front, right, back and left. 
Participants are asked to choose the reconstructed \threeD shape that better represents the human in the given color image.
Each participant is given $100$ samples to evaluate.
To teach participants, and to filter out the ones that do not understand the task, we set up $1$ tutorial sample, followed by $10$ warm-up samples, and then the evaluation samples along with catch trial samples inserted every $10$ evaluation samples.
Each catch trial sample shows a color image along with either 
(1) the reconstruction of a baseline method for this image and the \groundtruth scan that was rendered to create this image, or 
(2) the reconstruction of a baseline method for this image and the reconstruction for a different image (false positive), see \cref{fig:perceptual_study:catch_trial}.
Only participants that pass $70\%$ out of $10$ catch trials are considered. 
This leads to $28$ valid participants out of $36$ ones. 
Results are reported in \cref{tab: perceptual study}.

\begin{figure*}[t]
    \centering
    \begin{subfigure}{\linewidth}
        \includegraphics[width=0.5\linewidth]{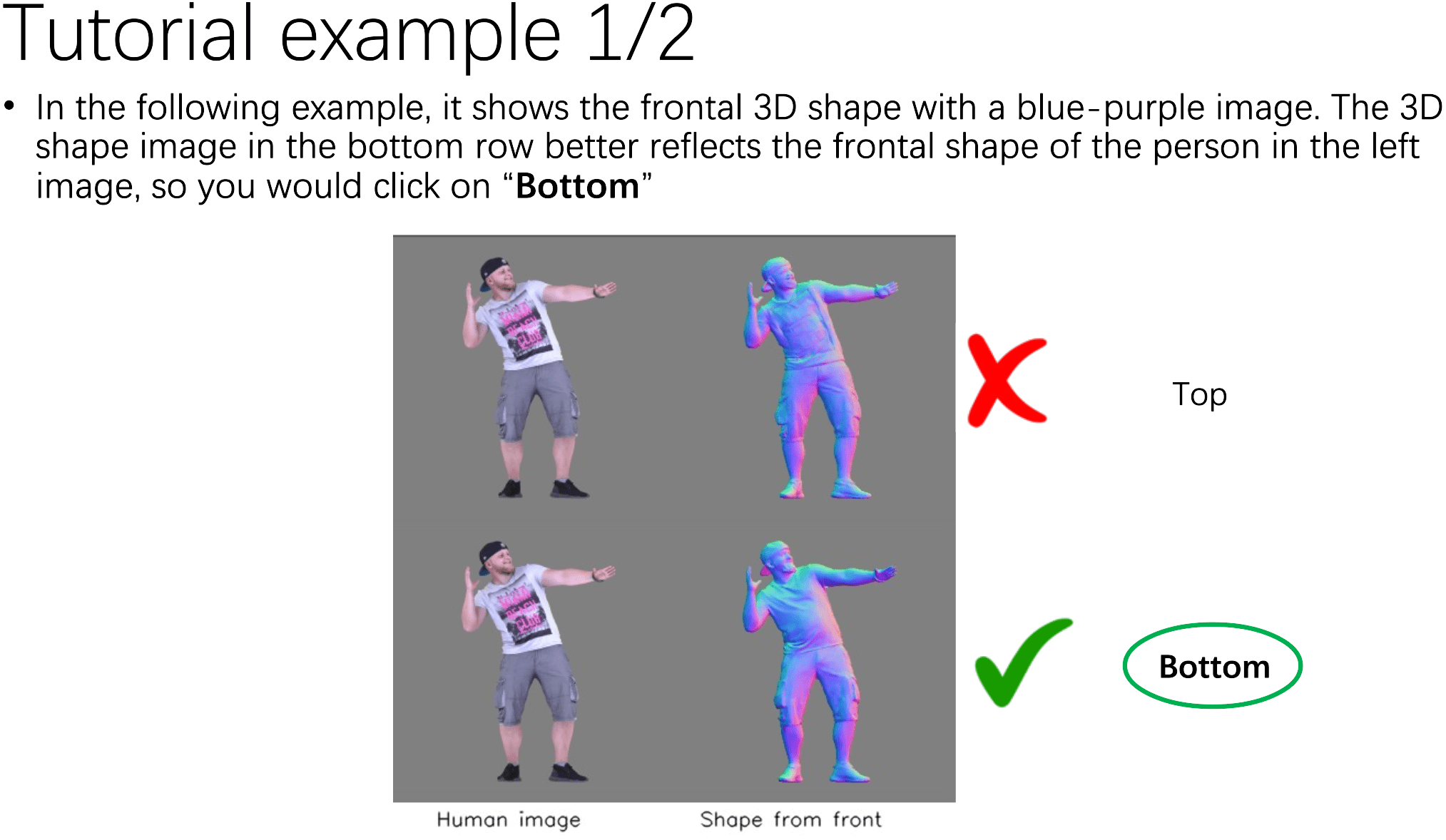}
        \includegraphics[width=0.5\linewidth]{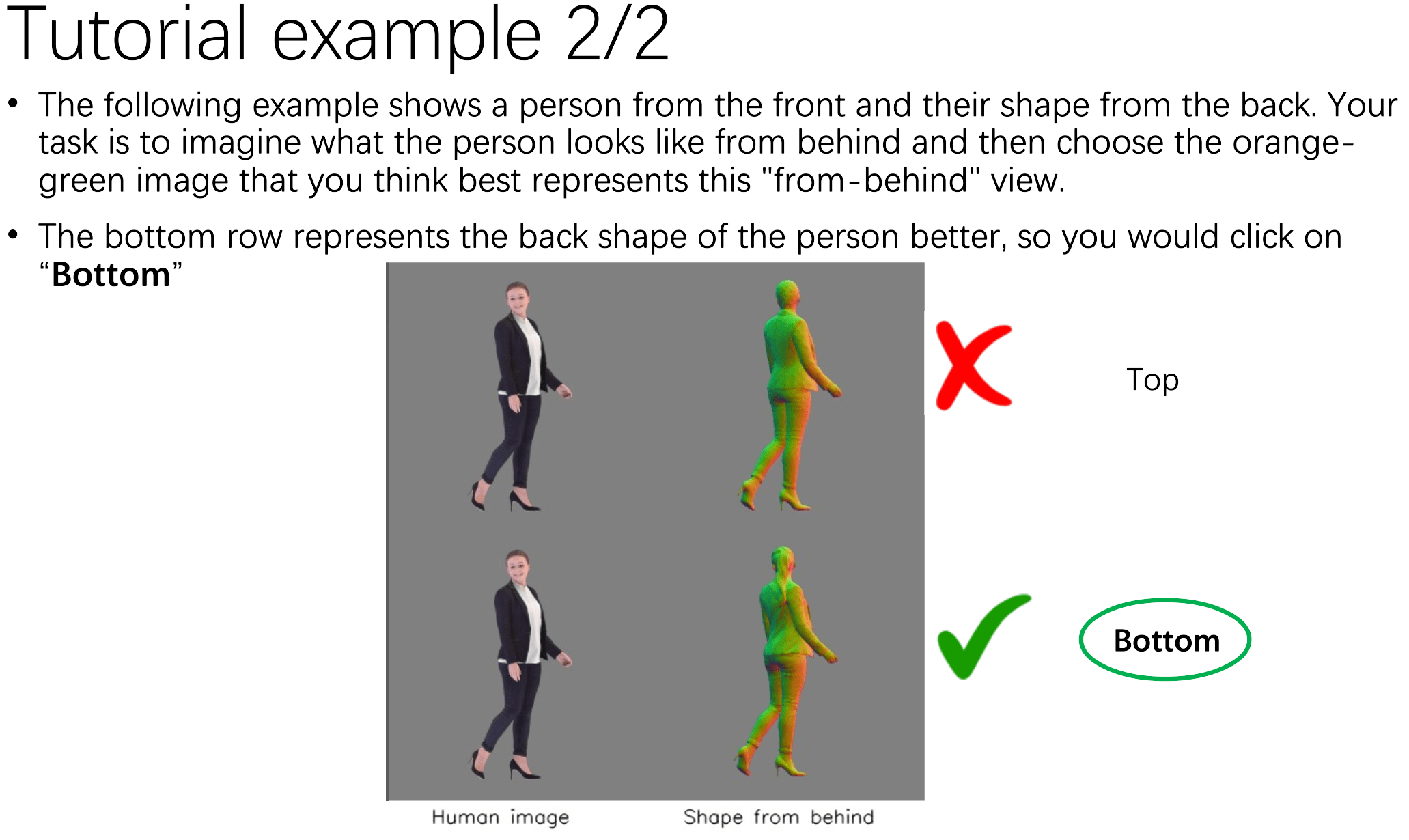}
        \caption{The two tutorial samples.}
    \label{fig:perceptual_study_normal:tutorial}
    \end{subfigure}
    \begin{subfigure}{0.49\linewidth}
        \includegraphics[width=0.49\linewidth]{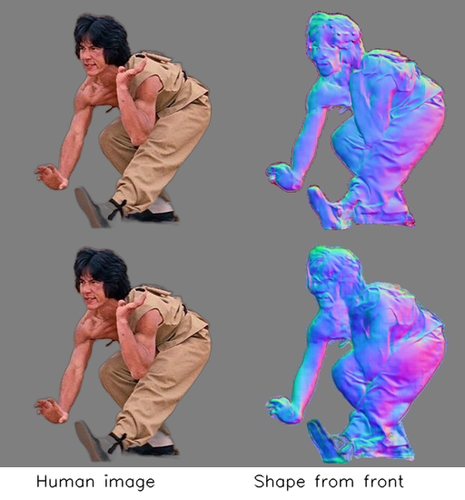} 
        \hfill
        \includegraphics[width=0.49\linewidth]{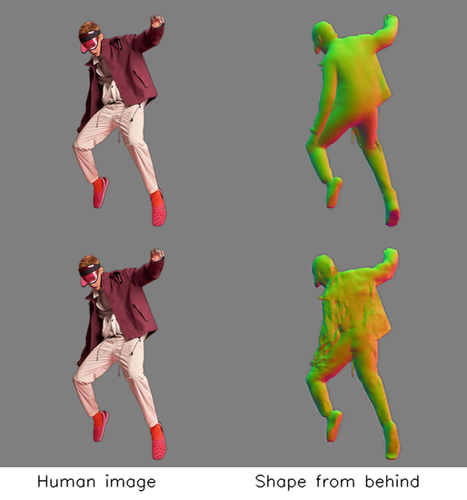} 
        \caption{Two evaluation samples.}
    \label{fig:perceptual_study_normal:evaluation}
    \end{subfigure}
    \hfill
    \begin{subfigure}{0.49\linewidth}
        \includegraphics[width=0.49\linewidth]{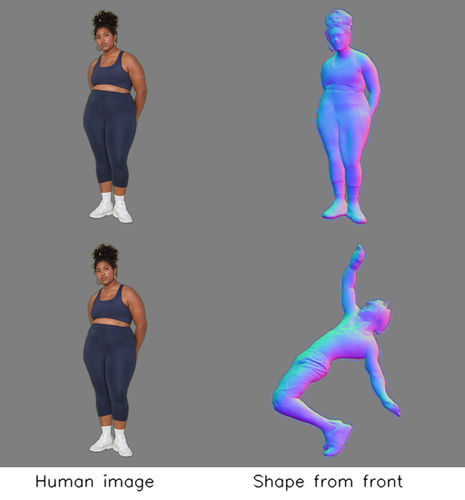}
        \hfill
        \includegraphics[width=0.49\linewidth]{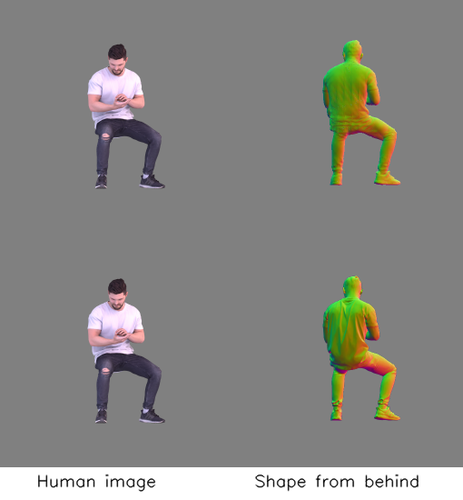} 
        \caption{Two catch trial samples.}
    \label{fig:perceptual_study_normal:catch_trial}
    \end{subfigure}
    \caption{Some samples in the perceptual study to evaluate the effect of the \textbf{body prior} for \textbf{normal prediction} on \inthewild images.}
    \label{fig:perceptual_study_normal}
\end{figure*}

\begin{figure*}[t]
    \centering
    \begin{subfigure}{\linewidth}
    \includegraphics[trim=000mm 000mm 000mm 000mm, clip=True, width=1.00 \linewidth]{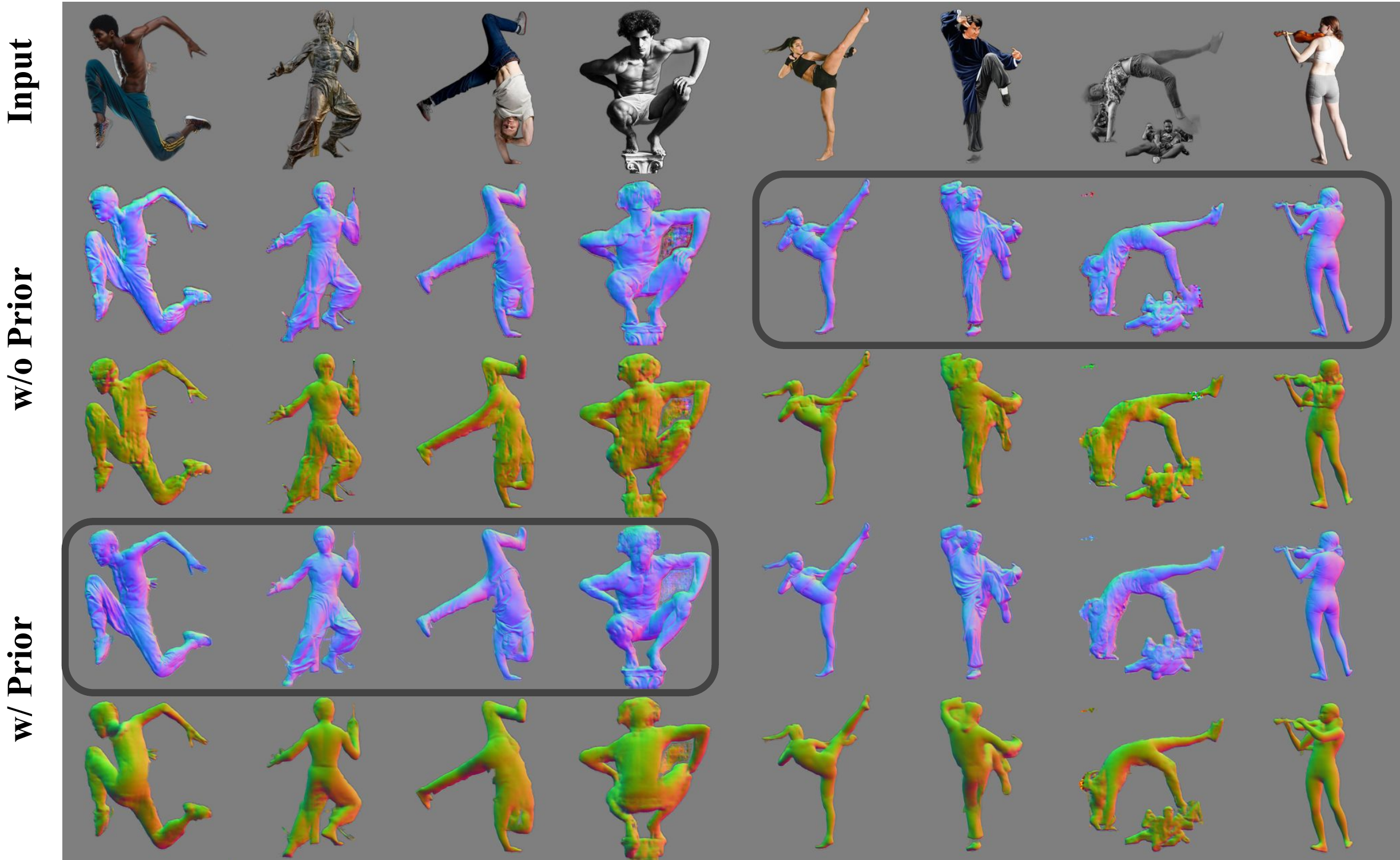}
    \caption{Examples of perceptual preference on \textbf{front} normal maps. Unanimously preferred results are in \Ovalbox{black boxes}. The back normal maps are for reference.\\}
    \vspace{-0.5em}
    \label{fig:normal_perceptual:front}
    \end{subfigure}
    \begin{subfigure}{\linewidth}
    \includegraphics[width=1.00 \linewidth]{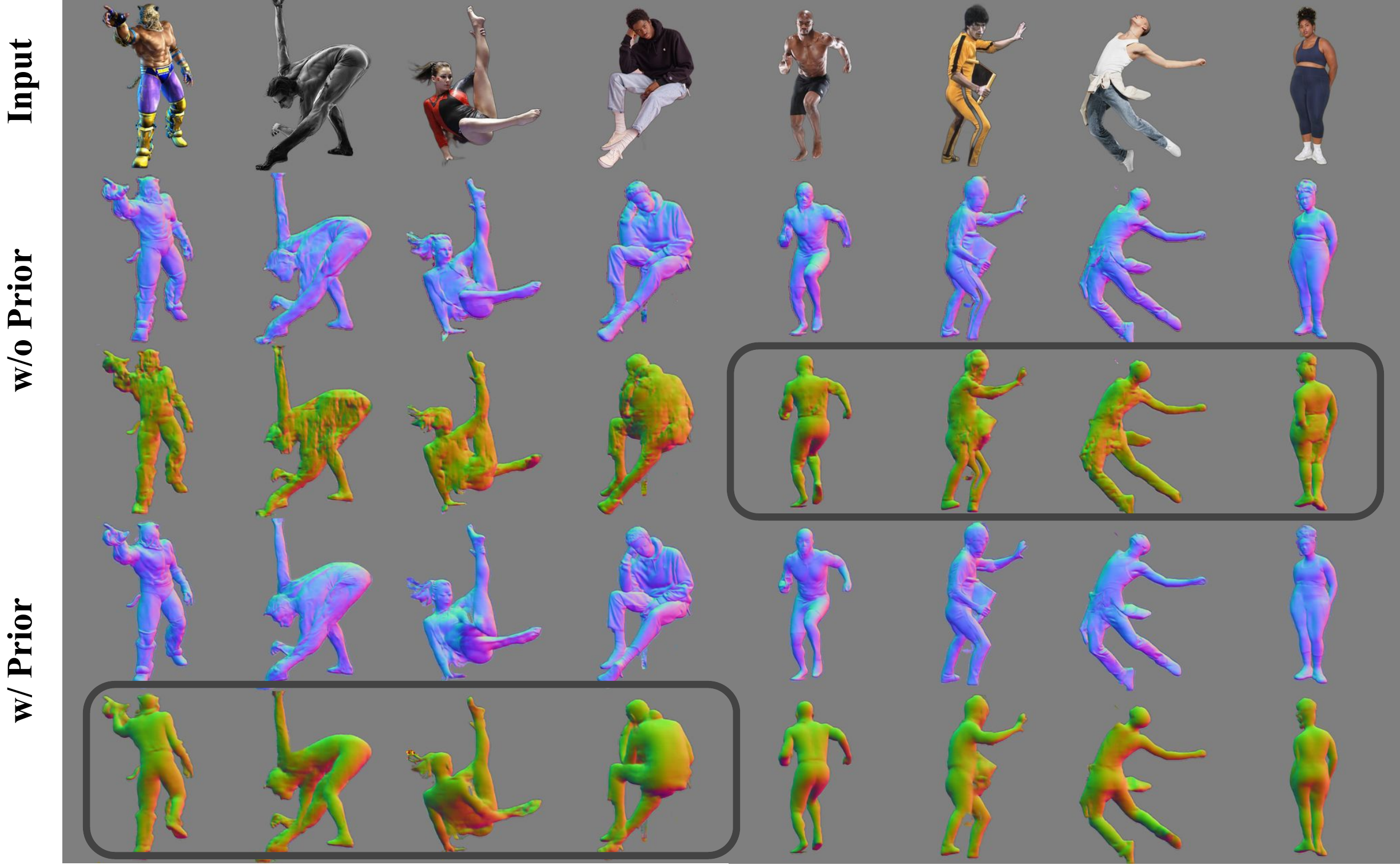}    
    \caption{Examples of perceptual preference on \textbf{back} normal maps. Unanimously preferred results are in \Ovalbox{black boxes}. The front normal maps are for reference.}
    \vspace{-0.5em}
    \label{fig:normal_perceptual:back}
  \end{subfigure}
  \caption{Qualitative results to evaluate the effect of body prior for normal prediction on \inthewild images.}
  \vspace{-0.5em}
  \label{fig:normal_perceptual}
\end{figure*}

\smallskip
\qheading{Normal map prediction.} 
To evaluate the effect of the body prior for normal map prediction on \inthewild images, we conduct a perceptual study against prediction without the body prior. 
We use \ac{amt}, and show participants a color image along with a pair of predicted normal maps from two methods. 
Participants are asked to pick the normal map that better represents the human in the image. 
Front- and back-side normal maps are evaluated separately. 
See \cref{fig:perceptual_study_normal} for some samples. 
We set up $2$ tutorial samples, 10 warm-up samples, $100$ evaluation samples and $10$ catch trials for each subject.
The catch trials lead to $20$ valid subjects out of $24$ participants. 
We report the statistical results in \cref{tab: suppl perceptual study norm}. 
A chi-squared test is performed with a null hypothesis that the body prior does not have any influence. 
We show some results in \cref{fig:normal_perceptual}, where all participants unanimously prefer one method over the other. 
While results of both methods look generally similar on front-side normal maps, using the body prior usually leads to better back-side normal maps.

\begin{table}[ht]
\centering
\footnotesize
\begin{tabular}{c|ccc}
 & w/ \smpl prior & w/o \smpl prior & P-value\\
\shline
Preference (front) & 47.3\% & 52.7\% & 8.77e-2\\
Preference (back) & 52.9\% & 47.1\% & 6.66e-2
\end{tabular}
\caption{Perceptual study on normal prediction.}
\label{tab: suppl perceptual study norm}
\end{table}

\begin{table}[t]
\centering
\footnotesize
\begin{tabular}{l|c|ccc}
& w/ global & pixel & point &  total \\
& encoder & dims & dims &  dims \\
\shline
\pifuSIM & \cmark & 12 & 1 & 13\\
\pamirSIM & \cmark & 6 & 7 & 13\\
$\text{\modelname}_\text{enc(I,N)}$ & \cmark &  6 & 7 & 13\\
$\text{\modelname}_\text{enc(N)}$ & \cmark & 6 & 7 & 13\\
$\text{\modelname}$ & \xmark & 0 & 7 & 7\\
\end{tabular}
\caption{Feature dimensions for various approaches. 
``pixel dims'' and ``point dims'' denote the feature dimensions encoded from pixels (image/normal maps) and \threeD body prior, respectively.}
\label{tab:feat_dim}
\end{table}

\subsection{Implementation details (\cref{sec: baseline models})}
    
\qheading{Network architecture.}
Our body-guided normal prediction network uses the same architecture as \pifuhd \cite{saito2020pifuhd}, originally proposed in \cite{justin2016perceploss}, and consisting of residual blocks with $4$ down-sampling layers.
The image encoder for \pifuSIM, \pamirSIM, and $\text{\modelname}_\text{enc}$ is a stacked hourglass  \cite{newell2016hourglass} with $2$ stacks, modified according to \cite{aaron2018volreg}. 
\cref{tab:feat_dim} lists feature dimensions for various methods; ``total dims'' is the neuron number for the first MLP layer (input). 
The number of neurons in each MLP layer is: $13$ ($7$ for \modelname), $512$, $256$, $128$, and $1$, with skip connections at the $3$rd, $4$th, and $5$th layers. 

\medskip
\qheading{Training details.}
For training $\normNet$ we do not use \thuman due to its low-quality texture (see \cref{tab:dataset}). On the contrary, $\imFunc$ is trained on both \agora and \thuman. The front-side and back-side normal prediction networks are trained individually with batch size of $12$ under the objective function defined in \cref{eq:normal-loss}, where we set $\lambda_\text{VGG} = 5.0$. 
We use the ADAM optimizer with a learning rate of $1.0 \times 10^{-4}$ until convergence at $80$ epochs. 


\medskip
\qheading{Test-time details.}
During inference, to iteratively refine \smpl and the predicted clothed-body normal maps, we perform $50$ iterations \highlight{(each iteration takes $\sim460$ ms on a Quadro RTX $5000$ GPU)} and set $\lambda_\text{N} = 2.0$ in \cref{eq:body-fit}. 
\highlight{We conduct an experiment to show the influence of the number of iterations (\#iterations) on accuracy, see \cref{tab:loop}.}

The resolution of the queried occupancy space is $256^3$. 
We use \specific{rembg}\footnote{\url{https://github.com/danielgatis/rembg}} to segment the humans in \inthewild images, and use \specific{Kaolin}\footnote{\url{https://github.com/NVIDIAGameWorks/kaolin}} to compute per-point the signed distance, $\bodySDF$, and barycentric surface normal, $\bodyNormFeat$.

\begin{table}
\vspace{-0.2 em}
\hspace{-0.5 em}
\resizebox{0.5\linewidth}{!}{
\begin{minipage}{5.5cm}
\setlength{\parindent}{0em}
\addtolength{\tabcolsep}{-3pt}
\centering
\begin{tabular}{c|ccc}
    \multicolumn{1}{c|}{\# iters (460ms/it)} & 0 & 10 & 50\\
    \shline
    Chamfer $\downarrow$ & 1.417 & 1.413 & \textbf{1.339}\\
    P2S     $\downarrow$ & 1.436 & 1.515 & \textbf{1.378}\\
    Normal  $\downarrow$ & 0.082 & 0.077 & \textbf{0.074}\\
\end{tabular}
\caption{\modelname errors \wrt iterations}
\label{tab:loop}
\end{minipage}}
\hspace{-0.2 em}
\resizebox{0.5\linewidth}{!}{
\begin{minipage}{5.5cm}
\setlength{\parindent}{0em}
\addtolength{\tabcolsep}{-3pt}
\centering
\begin{tabular}{c|ccc}
    \multicolumn{1}{c|}{Receptive field} & 139 & 271 & 403\\
    \shline
    Chamfer $\downarrow$ & 1.418 & 1.478 & \textbf{1.366}\\
    P2S     $\downarrow$ & 1.236 & 1.320 & \textbf{1.214}\\
    Normal  $\downarrow$ & 0.083 & 0.084 & \textbf{0.078}\\
\end{tabular}
\caption{\pamir's receptive field}
\label{tab:rp}
\end{minipage}}
\vspace{-1.5 em} 
\end{table}

\highlight{
\medskip
\qheading{Discussion on receptive field size.}
As \cref{tab:rp} shows, simply reducing the size of receptive field of \pamir does not lead to better performance. This shows that our informative 3D features as in \cref{eq:point feature} and normal maps $\predCloNormImg$ also play important roles for robust reconstruction. A more sophisticated design of smaller receptive field may lead to better performance and we would leave it for future research.
}

\newpage

\section{More Quantitative Results (\cref{sec: evaluation})}
\Cref{tab:sup-benchmark} compares several \modelname variants conditioned on perturbed \smplx meshes. 
For the plot of \cref{fig:datasize} of the main paper (reconstruction error \wrt training-data size), extended quantitative results are shown in \cref{tab:geo-datasize}.

\begin{table}[h]
\centering{
\resizebox{\linewidth}{!}{
 \begin{tabular}{lc|ccccc}
    \multicolumn{2}{c|}{Training set scale} & 1/8x & 1/4x & 1/2x & 1x & 8x\\
    \shline
    \multirow{2}{*}{\pifuSIM } & Chamfer $\downarrow$ & 3.339 & 2.968 & 2.932 & 2.682 & 1.760 \\
    & P2S $\downarrow$ & 3.280 & 2.859 & 2.812 & 2.658 & 1.547 \\
    \multirow{2}{*}{\pamirSIM } & Chamfer $\downarrow$ & 2.024 & 1.780 & 1.479 & 1.350 & \colorbox{LightCyan}{1.095} \\
    & P2S $\downarrow$ & 1.791 & 1.778 & 1.662 & 1.283 & 1.131 \\
    \multirow{2}{*}{\modelname} & Chamfer $\downarrow$ & 1.336 & 1.266 & 1.219 & \colorbox{LightCyan}{1.142} & \colorbox{LightCyan}{\textbf{1.036}}\\
    & P2S $\downarrow$ & 1.286 & 1.235 & 1.184 & \colorbox{LightCyan}{1.065} & \colorbox{LightCyan}{\textbf{1.063}}\\
 \end{tabular}}}
\vspace{-0.5 em}
\caption{Reconstruction error (cm) \wrt training-data size. 
``Training set scale'' is defined as the ratio \wrt the $450$ scans used in \cite{saito2019pifu,saito2020pifuhd}. 
The ``8x'' setting is all $3,709$ scans of \agora~\cite{patel2021agora} and \thuman~\cite{zheng2019deephuman}.  
\colorbox{LightCyan}{Results} outperform \groundtruth \smplx, which has 1.158 cm and 1.125 cm for Chamfer and P2S in \cref{tab:benchmark}.}
\label{tab:geo-datasize}
\end{table}

\section{More Qualitative Results (\cref{sec: applications})}
\begin{figure*}
    \centering
    \includegraphics[width=1.04 \linewidth]{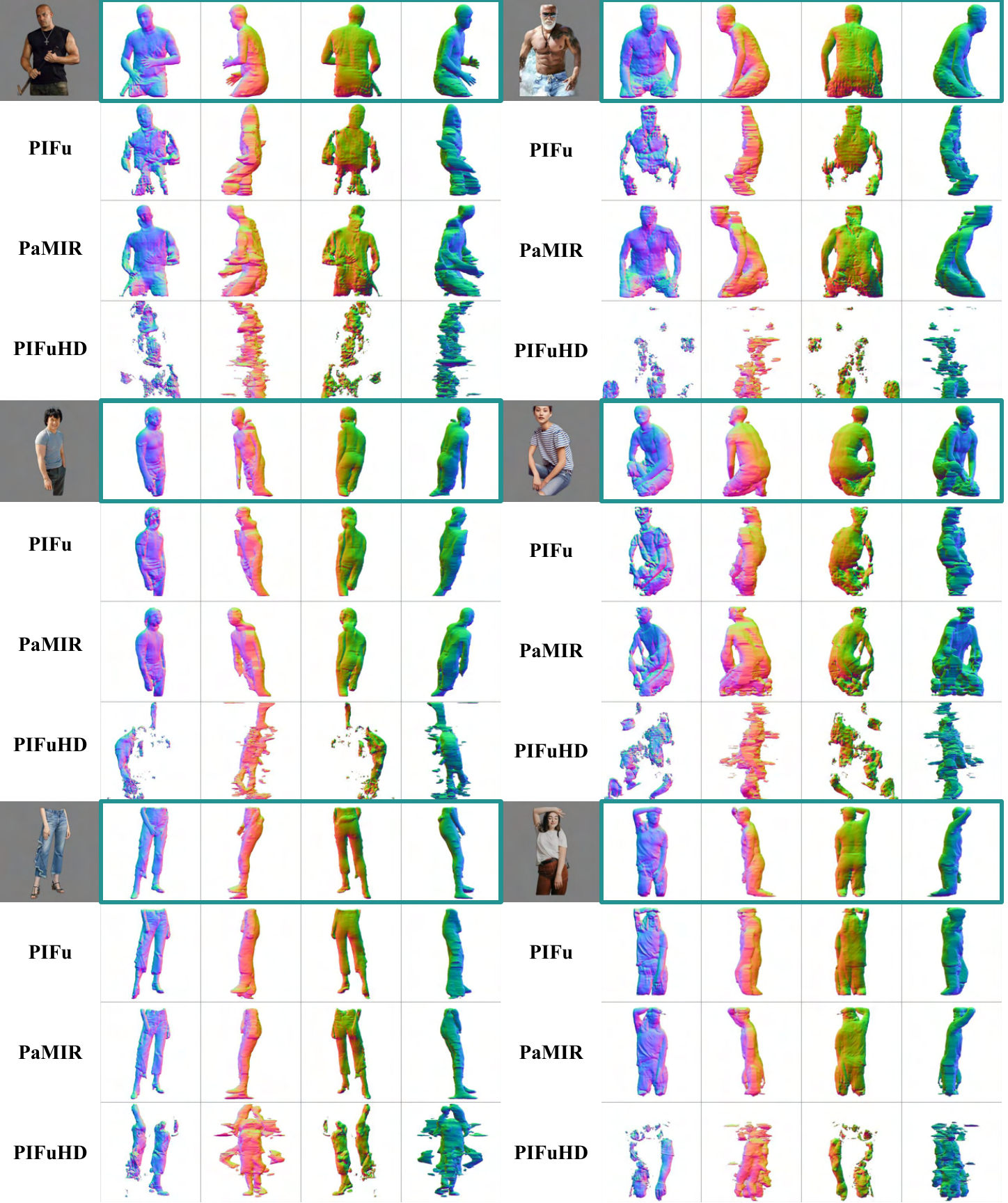}
    \caption{Qualitative comparison (\textcolor{GreenColor}{\modelname} vs \sota) on images with out-of-frame cropping.}
    \label{fig:out-of-frame}
\end{figure*}

\begin{figure*}
\centering
    \includegraphics[width=1.04 \linewidth]{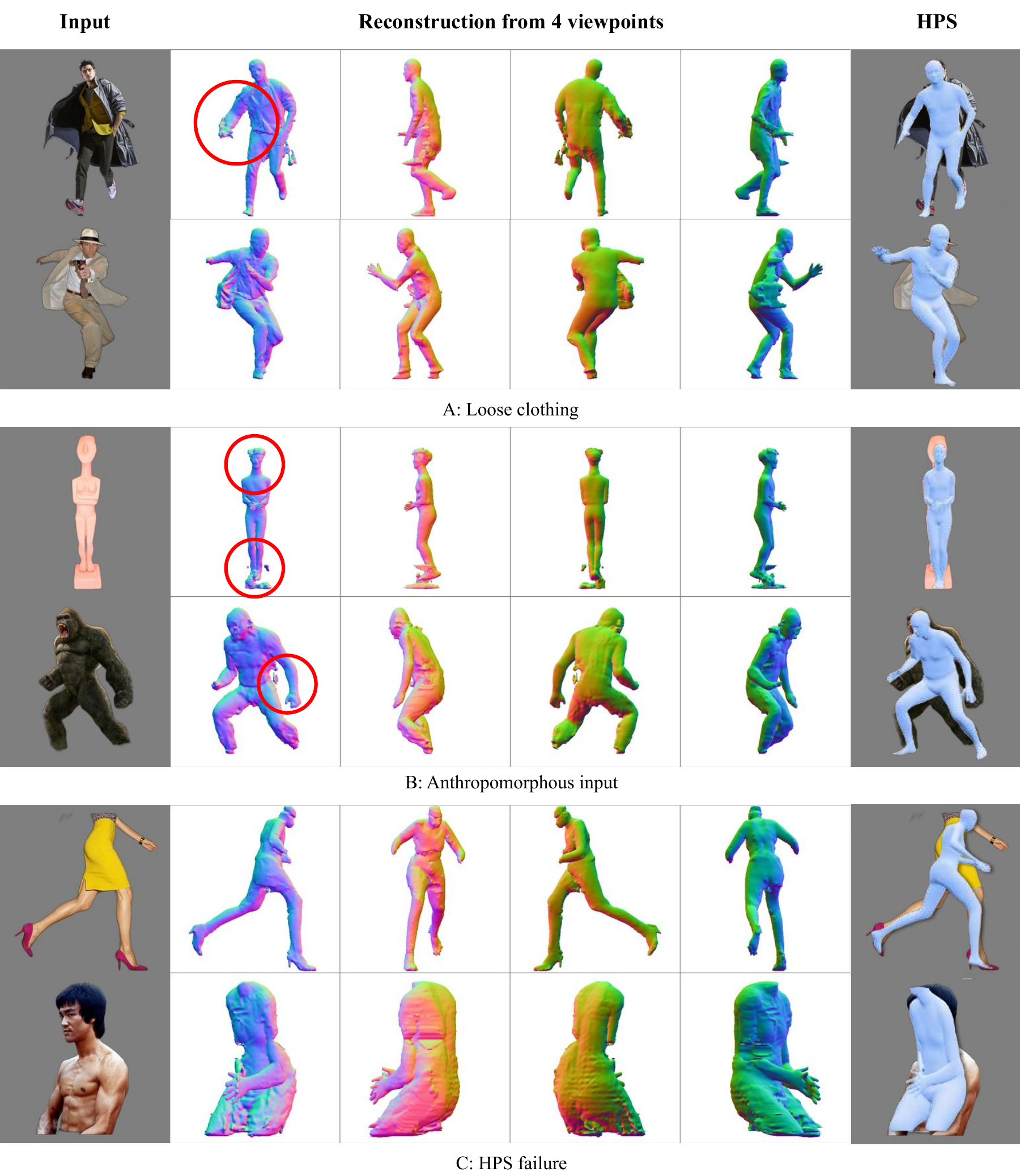}
    \caption{More failure cases of \modelname.}
    \label{fig:failure cases}
\end{figure*}

\Cref{fig:comparison1,fig:comparison2,fig:comparison3} show reconstructions for \inthewild images, rendered from four different view points; normals are color coded.
\Cref{fig:out-of-frame} shows reconstructions for images with \oof cropping.
\Cref{fig:failure cases} shows additional representative failures.
The \video on our website shows animation examples created with \modelname and \scanimate. 


\newcommand{\comparisonCaptionSupMat}{Qualitative comparison of reconstruction for \textcolor{GreenColor}{\modelname} vs \sota. Four view points are shown per result.}

\begin{figure*}
    \centering
    \vspace{-2.0 em}
    \hspace{-3.0 em}
    \includegraphics[width=1.07 \textwidth]{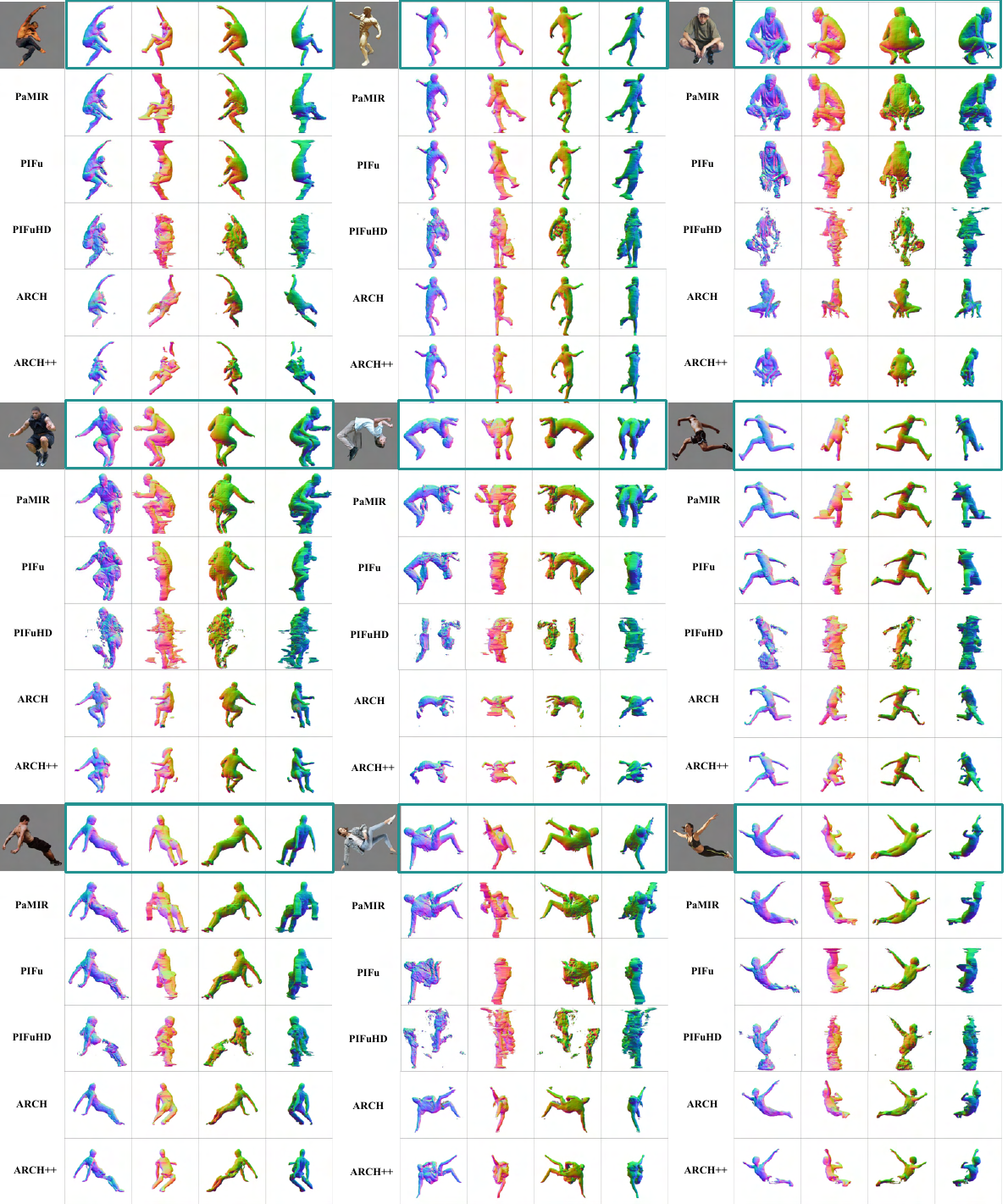}
    \caption{\comparisonCaptionSupMat}
    \label{fig:comparison1}
\end{figure*}

\begin{figure*}
    \centering
    \vspace{-2.0 em}
    \hspace{-3.0 em}
    \includegraphics[width=1.07 \textwidth]{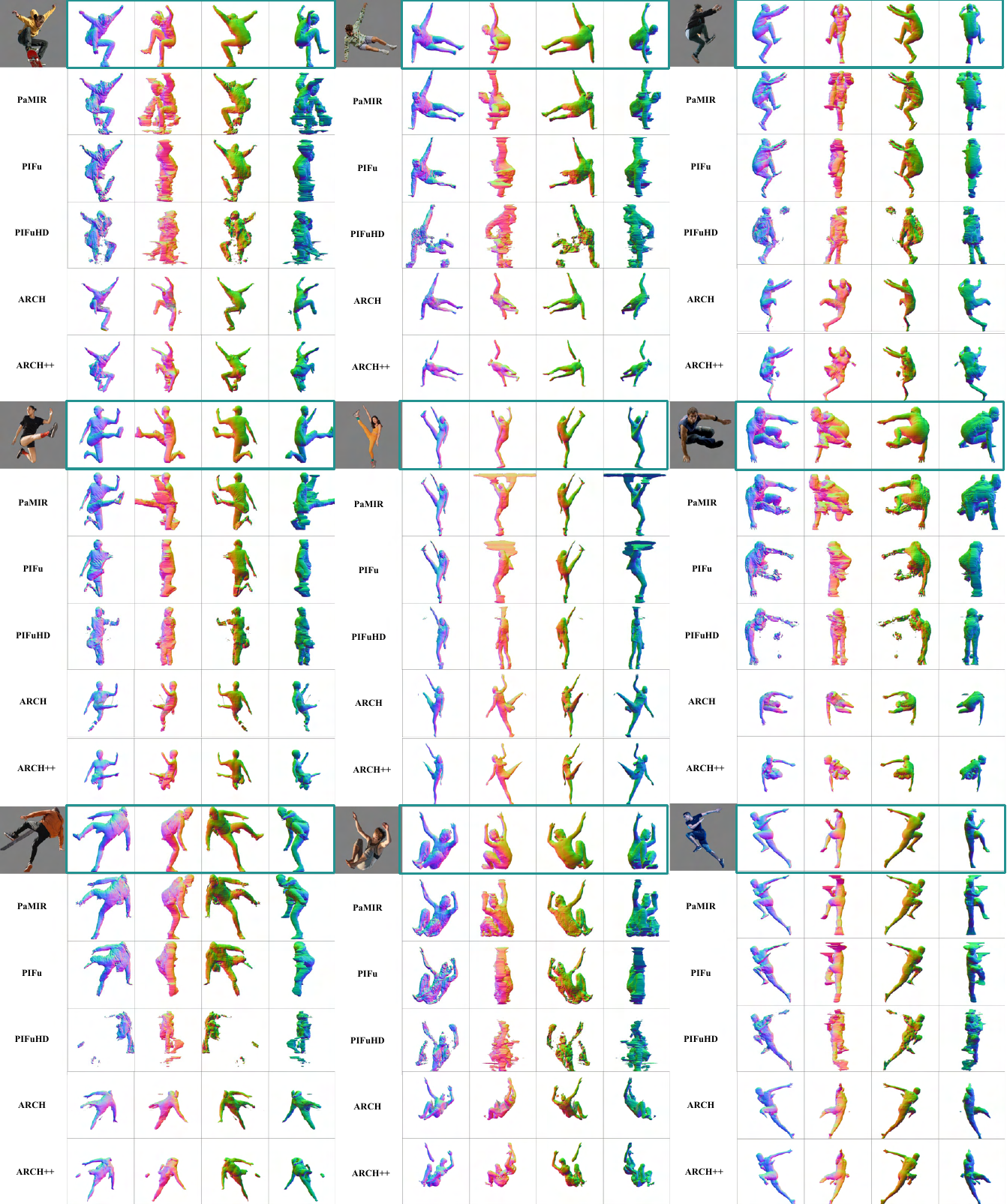}
    \caption{\comparisonCaptionSupMat}
    \label{fig:comparison2}
\end{figure*}

\begin{figure*}
    \centering
    \vspace{-2.0 em}
    \hspace{-3.0 em}
    \includegraphics[width=1.07 \textwidth]{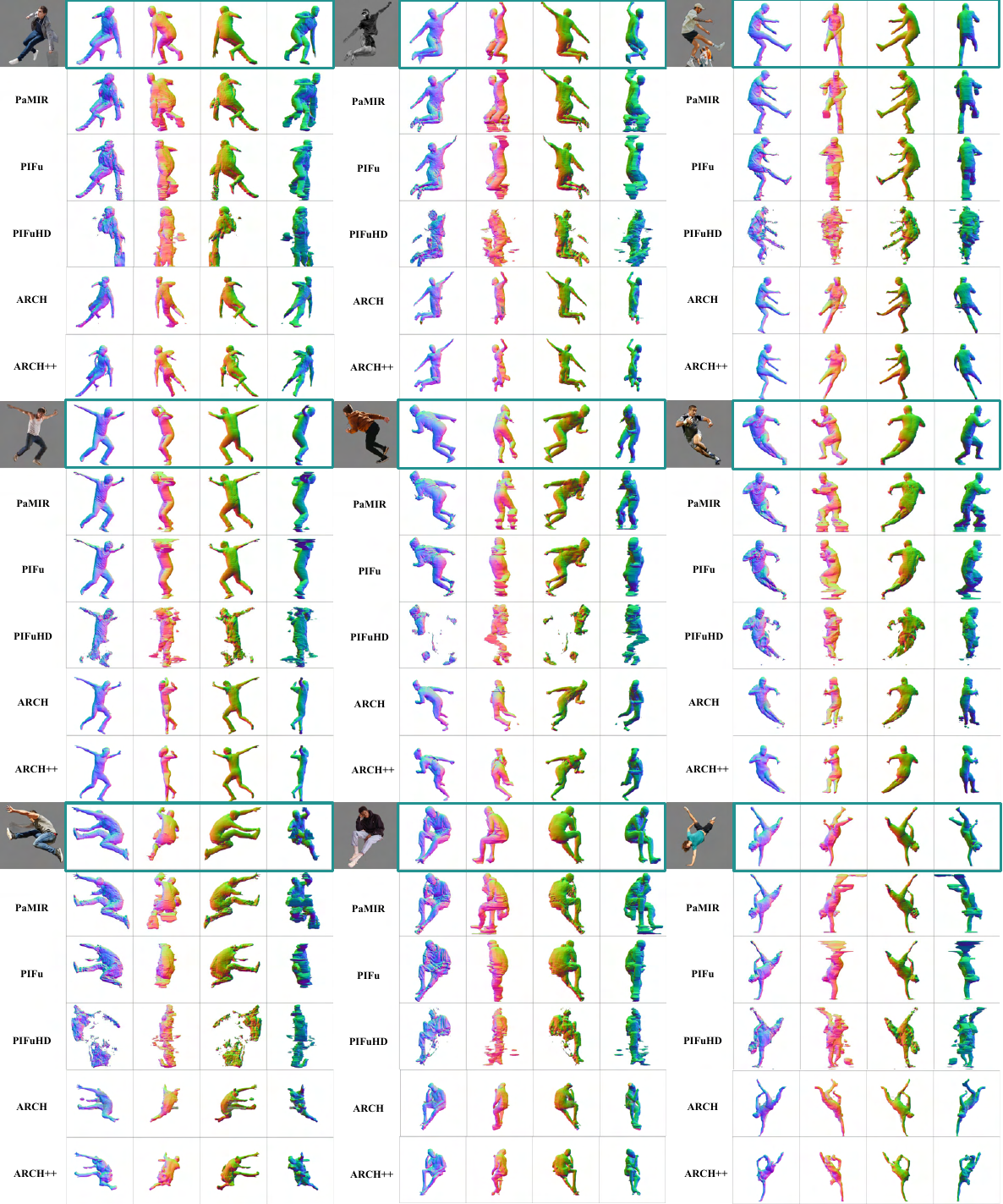}
    \caption{\comparisonCaptionSupMat}
    \label{fig:comparison3}
\end{figure*}

\end{appendices}


\clearpage
{\small
\balance
\bibliographystyle{config/ieee_fullname}
\bibliography{references}
}


\end{document}